\documentclass{article}
\pdfoutput=1

\usepackage{graphicx}
\usepackage{cite}
\usepackage{amsmath}

\newcommand{\widefigwidth}{5.0in}
\newcommand{\figwidth}{3.5in}
\newcommand{\mfigwidth}{3.0in}
\newcommand{\smallfigwidth}{2in}

\newcommand{\eq}[1]{Eq.~(\ref{eq.#1})} 
\newcommand{\fig}[1]{Fig.~\ref{fig.#1}}

\newcommand{\tbl}[1]{Table~\ref{table.#1}}

\newcommand{\sect}[1]{Section~\ref{sect.#1}}
\newcommand{\sectA}[1]{Appendix~\ref{sect.#1}}
\newcommand{\sectlabel}[1]{\label{sect.#1}}
\newcommand{\eqlabel}[1]{\label{eq.#1}}
\newcommand{\figlabel}[1]{\label{fig.#1}}
\newcommand{\tbllabel}[1]{\label{table.#1}}

\newcommand{\BoltzmannConstant}{\ensuremath{k_{\text{B}}}}

\newcommand{\invivo}{\textit{in vivo}}

\renewcommand\vec{\mathbf}

\newcommand{\angleTExtreme}{\ensuremath{\theta_{\textnormal{extreme}}}}
\newcommand{\angleToWall}{\ensuremath{\theta_{\textnormal{wall}}}} 
\newcommand{\angleXToWall}{\ensuremath{\phi_{\textnormal{wall}}}} 
\newcommand{\orientation}{\ensuremath{\psi}}

\newcommand{\density}{\ensuremath{\rho}}
\newcommand{\viscosity}{\ensuremath{\eta}}
\newcommand{\viscosityKinematic}{\ensuremath{\nu}}  
\newcommand{\Tfluid}{\ensuremath{T_{\textnormal{fluid}}}}

\newcommand{\vRobot}{\ensuremath{v_{\textnormal{robot}}}} 
\newcommand{\omegaRobot}{\ensuremath{\omega_{\textnormal{robot}}}} 
\newcommand{\speedRatio}{\ensuremath{s_{\textnormal{ratio}}}} 
\newcommand{\surfaceFraction}{\ensuremath{\lambda}}

\newcommand\Rey{\mbox{Re}}  
\newcommand\Pec{\mbox{Pe}}  
\newcommand\Wom{\mbox{Wo}}  
\newcommand\relPos{\textnormal{r.p.}}

\newcommand{\RMS}{\textsc{rms}}
\newcommand{\SNR}{\textsc{snr}}

\newcommand{\meter}{\mbox{m}}

\newcommand{\micron}{\mbox{$\mu$m}}
\newcommand{\nanometer}{\mbox{nm}}
\newcommand{\second}{\mbox{s}}
\newcommand{\millisecond}{\mbox{ms}}
\newcommand{\nanosecond}{\mbox{ns}}

\newcommand{\kilogram}{\mbox{kg}}

\newcommand{\pascal}{\mbox{Pa}}

\newcommand{\Kelvin}{\mbox{K}}
\newcommand{\radian}{\mbox{rad}}
\newcommand{\decibel}{\textnormal{dB}}

\title{Stress-Based Navigation for Microscopic Robots in Viscous Fluids}
\author{Tad Hogg\\
{\small Institute for Molecular Manufacturing}\\{\small Palo Alto, CA 94301}
}

\begin{document}
\maketitle

\begin{abstract}

Objects moving in fluids experience patterns of stress on their surfaces determined by their motion and the geometry of nearby boundaries.  Fish and underwater robots can use these patterns for navigation. This paper extends this stress-based navigation to microscopic robots in tiny vessels, where robots can exploit the physics of fluids at low Reynolds number. This applies, for instance, in vessels with sizes and flow speeds comparable to those of capillaries in biological tissues. We describe how a robot can use simple computations to estimate its motion, orientation and distance to nearby vessel walls from fluid-induced stresses on its surface. Numerically evaluating these estimates for a variety of vessel sizes and robot positions shows they are most accurate when robots are close to vessel walls.

\end{abstract}

\section{Introduction}

Microscopic robots have the potential to significantly improve biological research and medicine~\cite{li17,martel07,monroe09,schulz09},
by precisely targeting their activities to individual cells.
Fully realizing the benefits of such robots requires methods to guide them to specific targets. 

One method is human control. This is suitable when a robot's operator knows where the robot should go, can observe its location relative to the targets~\cite{medina-sanchez17}, and can either signal the robot where to go or exert forces to move it, e.g., with magnetic fields~\cite{dreyfus05,martel07a,rahmer17,yesin05}. 
However, these requirements are difficult to meet when robots operate deep within tissue, when targets are only recognizable by small-scale properties not accessible to external imaging, or when using many robots with different targets. 

Alternatively, robots could use autonomous navigation to find targets by themselves. For instance, robots that move until they bind to targets can work well when targets are uniquely identifiable by their surface characteristics, and when selective robot surfaces can be designed and fabricated. However, robots need not only rely on binding to recognize targets. Instead, robots can combine a variety of sensor information and use onboard computation to improve targeting accuracy.
This information could include chemical concentration~\cite{park08}, fluid shear~\cite{korin12}, light and temperature~\cite{sershen00}, as well as broadcast external signals for coarse localization or synchronizing activity~\cite{boehm09,freitas99}. 
As nanoscale sensors and the computational capacity of microscopic robots improve, so will the complexity of sensor information that robots can integrate to move toward and identify their targets. 

Realizing this possibility requires both sensors and computers small enough to fit within microscopic robots. While there are various proposed designs for such robots and their components~\cite{schulz09}, manufacturing the robots is a significant challenge. Nevertheless, recent progress on nanoscale sensors and molecular computing demonstrate the possibility of sufficiently small sensors and computers.
In particular, a variety of nanoscale sensors, with dimensions of $100\,\nanometer$ or less, have been demonstrated~\cite{arlett07,cullinan12,ekinci05}. Such devices can be extremely sensitive, even to interactions with single molecules. 
For computation, DNA nanorobots~\cite{douglas12} use a few logic operations to improve target recognition, and programmable microorganisms~\cite{ferber04} can perform some computation. Computation, power and sensors have been combined into complete devices at somewhat larger sizes than considered in this paper~\cite{koman18}.
Theoretical studies of more complex molecular computers~\cite{drexler92,merkle18} suggest that more capable computers could also be small enough to fit within microscopic robots of the sizes considered here.

For robots operating in vessels, fluid stresses and vessel geometry can provide useful guidance toward targets in specific microenvironments.
For example, the structure of capillaries and their fluid stresses differ among organs~\cite{augustin17}, and between tumors and healthy tissue~\cite{nagy09,jain14}.
More generally, fluid stresses provide information about the object's environment within about a body length~\cite{sichert09}, although extracting this information requires significant computation for complex fluid flows~\cite{bouffanais10}.
This stress-based navigation is effective for fish in murky water~\cite{bleckmann09} 
and for some underwater vehicles~\cite{yang06a,vollmayr14}.

This paper examines how stresses could aid autonomous navigation of microscopic robots in tiny vessels. This requires accounting for their significantly different physical environments than encountered by fish or larger robots. For instance, viscous rather than inertial forces dominate the flow, i.e., the flow occurs at low Reynolds number~\cite{dusenbery09,purcell77}. Moreover, stresses and velocities are linearly related in such flows~\cite{happel83,kim05}, which can simplify how robots extract information from stresses.
Additional physical effects increasingly important at small sizes are Brownian motion and thermal noise limiting sensor accuracy.

This paper evaluates the consequences of these physical effects for two questions: What information about the vessel geometry and robot's motion is available from the pattern of stress on the robot's surface? And how can a robot extract that information with limited computational resources? We focus on short-term navigation, over times during which a robot moves a distance equal to a few times its size.
Specifically, the next section describes how a robot moves with fluid in a vessel and the resulting stresses on its surface. The following two sections evaluate how a robot can use stress measurements to estimate its orientation and position in the vessel, and its motion through the vessel. The paper concludes with a discussion of applications and extensions of these methods.
The appendix provides details on the fluid behavior, estimation methods, thermal noise and generalizations to more complex situations.

\section{Robots Moving with Fluid in Vessels}

This paper considers spherical robots in straight vessels, as shown in \fig{vessel geometry}, with parameters given in \tbl{fluid parameters}.
This vessel geometry is comparable to that of small blood vessels such as capillaries. With radii of curvature of tens of microns~\cite{pawlik81}, such vessels are approximately straight over the distances considered here.  
We determine stresses on the robot surface by numerically evaluating the flow and robot motion in a segment of the vessel. 
Due to symmetry, the robot's velocity lies in the cross section plane shown in \fig{vessel geometry}b.
The vessel size and fluid speed allows simplifying the numerical evaluation by using two-dimensional quasi-static Stokes flow, as described in \sectA{flow}.

\begin{table}
\centering
\begin{tabular}{lcc}
temperature	&\Tfluid	&$310\,\Kelvin$\\
density		& $\density$	&$10^3 \,\kilogram/\meter^3$	\\
viscosity		& $\viscosity$	&$10^{-3}\,\pascal \cdot \second$	\\
kinematic viscosity		& $\viscosityKinematic =\viscosity/\density$ & $10^{-6}\,\meter^2/\second$ \\
vessel diameter	&$d$		& $5\text{--}10\,\micron$\\
maximum flow speed	& $u$	& $200\text{--}2000\,\micron/\second$\\
Reynolds number	&$\Rey=u d/\viscosityKinematic$	& $<0.04$\\ 
%
robot radius	& $r$	& $1\,\micron$\\
robot speed	&$\vRobot$	& $<u$\\
robot angular velocity	&$\omegaRobot$	&$<u/r$\\
\end{tabular}
\caption{Typical parameters for fluids and microscopic robots considered here. 
In some cases, the last column gives a range of values or approximate bounds rather than a single value. Values for the scenarios considered in this paper are within about a factor of two of these bounds. 
}\tbllabel{fluid parameters}
\tbllabel{robot parameters}
\end{table}

\begin{figure}
\centering
\begin{tabular}{cc}
\includegraphics[width=\mfigwidth]{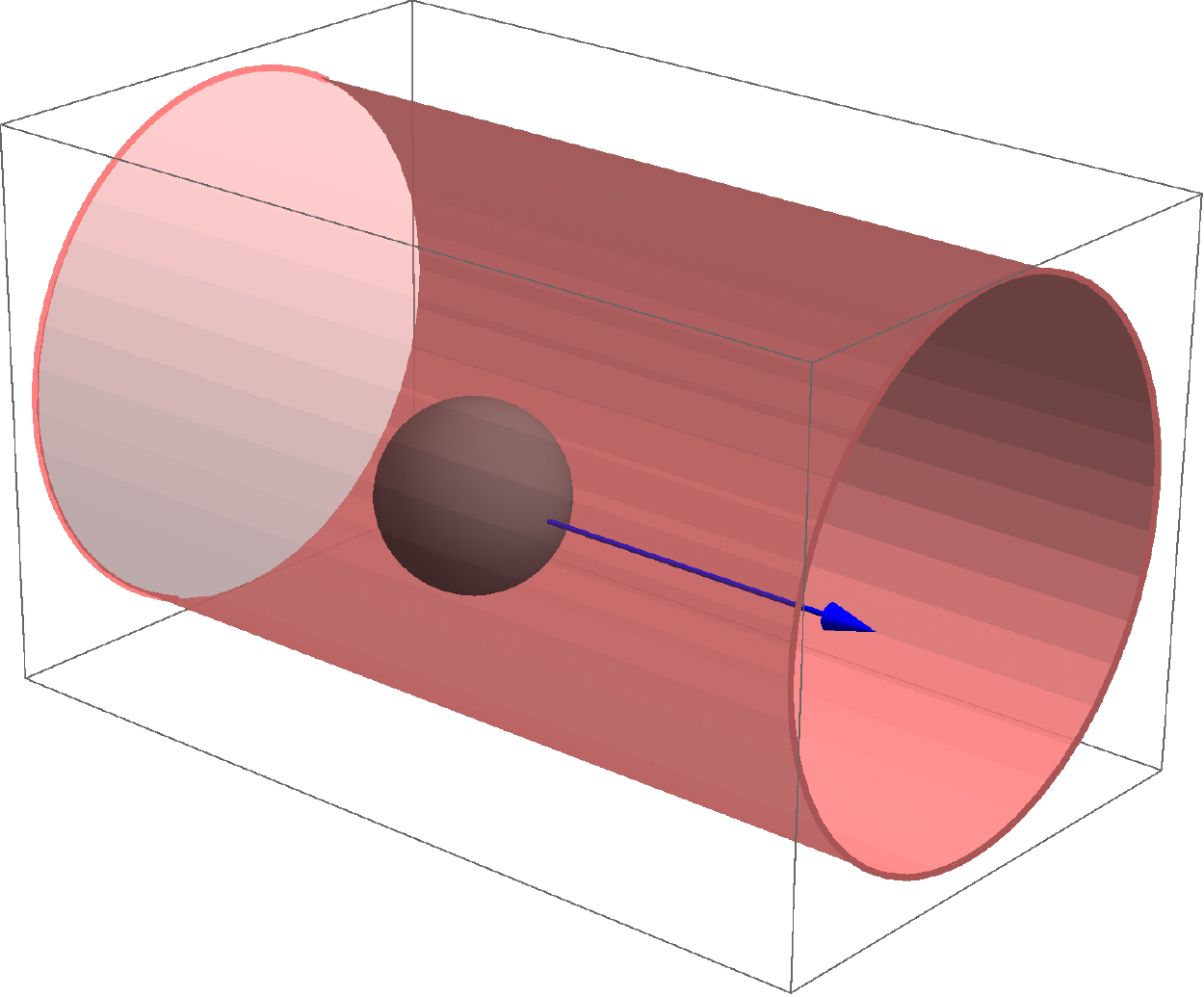}	&	\includegraphics[width=\mfigwidth]{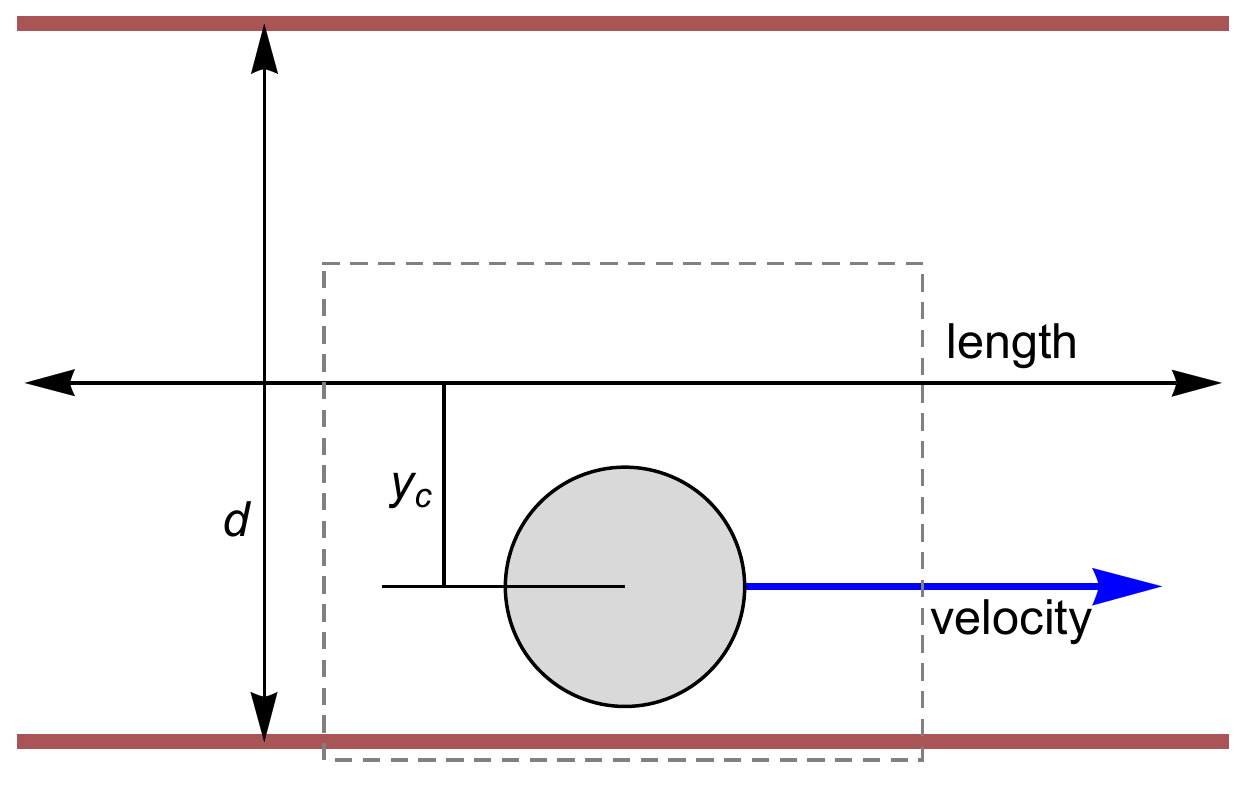} \\
(a) & (b) \\
\end{tabular}
\caption{Robot and vessel geometry. (a) A spherical robot in a cylindrical vessel. The blue arrow indicates the direction of the robot's motion. (b) Cross section of the vessel, with diameter $d$, and sphere, with radius $r$, whose center is at position $y_c$ with respect to the cylinder's axis, and distance $d/2-|y_c|$ from the vessel wall. This is also the geometry for the corresponding two-dimensional flow. The dashed rectangle is the part of the vessel shown in \fig{flow}.}\figlabel{vessel geometry}
\end{figure}

\begin{table}
\centering
\begin{tabular}{lcc}
robot position	&$y_c$	& $1.7\,\micron$\\
vessel diameter	&$d$		& $6\,\micron$\\
maximum inlet flow speed	&$u$		& $1000\,\micron/\second$\\
vessel segment length	&	&$10\,\micron$\\
\end{tabular}
\caption{Example parameters for geometry and fluid flow illustrated in \fig{vessel geometry}.}\tbllabel{example scenario}
\end{table}

\fig{flow} shows the flow and stresses on the robot surface using the parameters in \tbl{example scenario} and an arbitrary choice for the robot's orientation with respect to the vessel.
The fluid pushes the robot through the vessel at $530\,\micron/\second$ and with angular velocity $-150\,\radian/\second$, i.e., clockwise rotation.
\fig{Fourier}c shows how stresses vary over the robot surface. 
A useful representation of the stress pattern is its Fourier decomposition. In this example, the largest contribution is from the second mode, for both normal and tangential components of the stress, as shown in \fig{Fourier}d.

\begin{figure}
\centering 
\begin{tabular}{cc}
\includegraphics[width=2.5in]{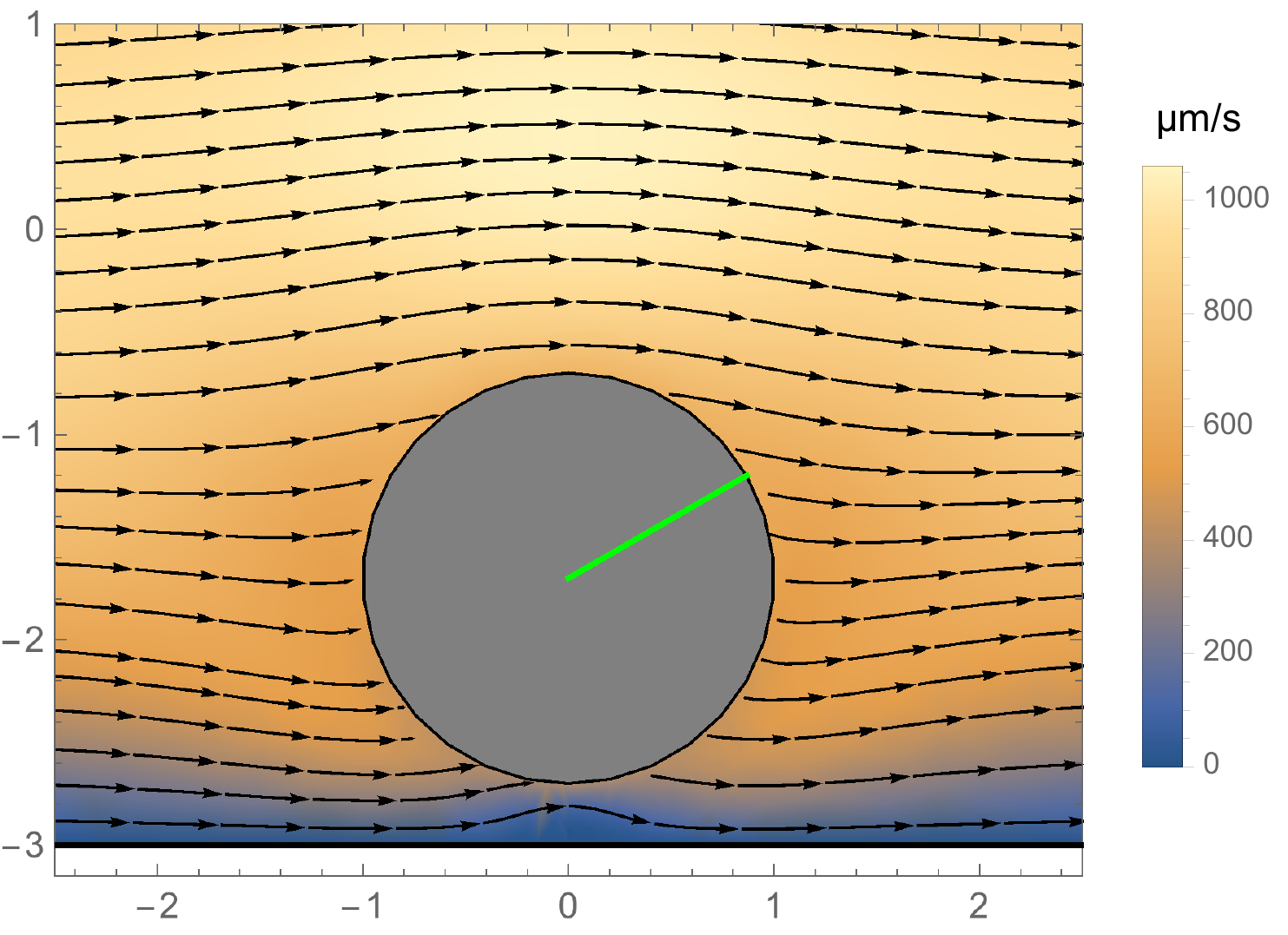} &
\includegraphics[width=2.5in]{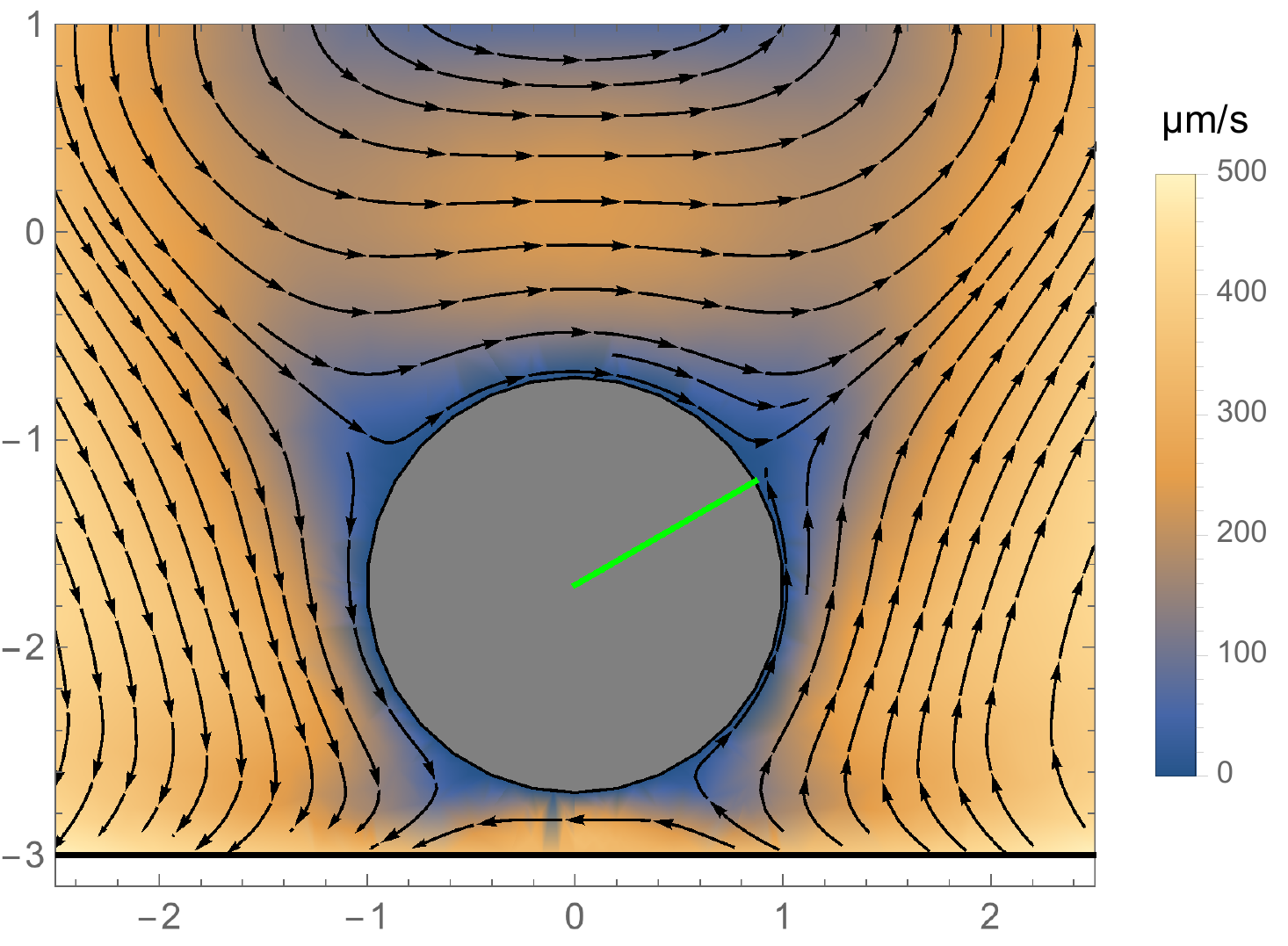}\\
(a) & (b)\\ \hline
\includegraphics[width=1.8in]{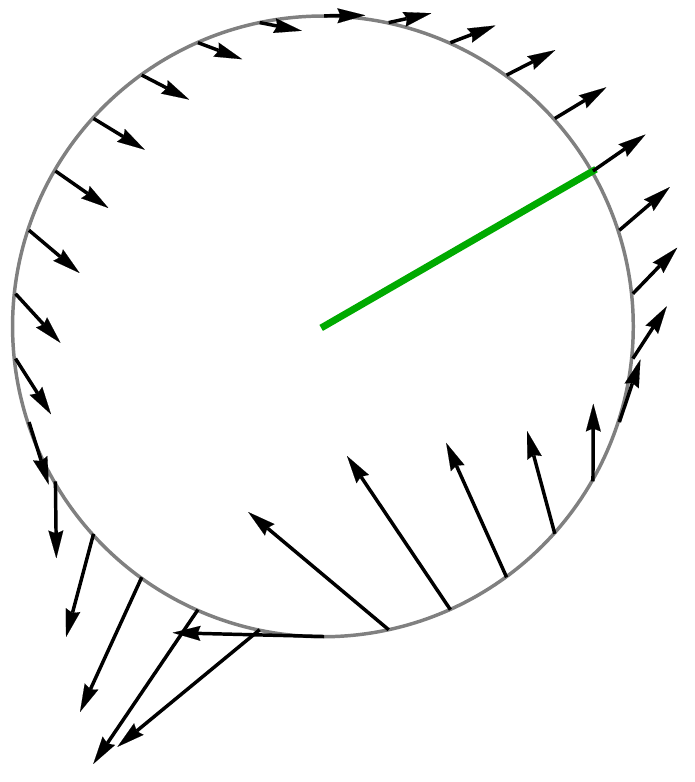} &
\includegraphics[width=2.8in]{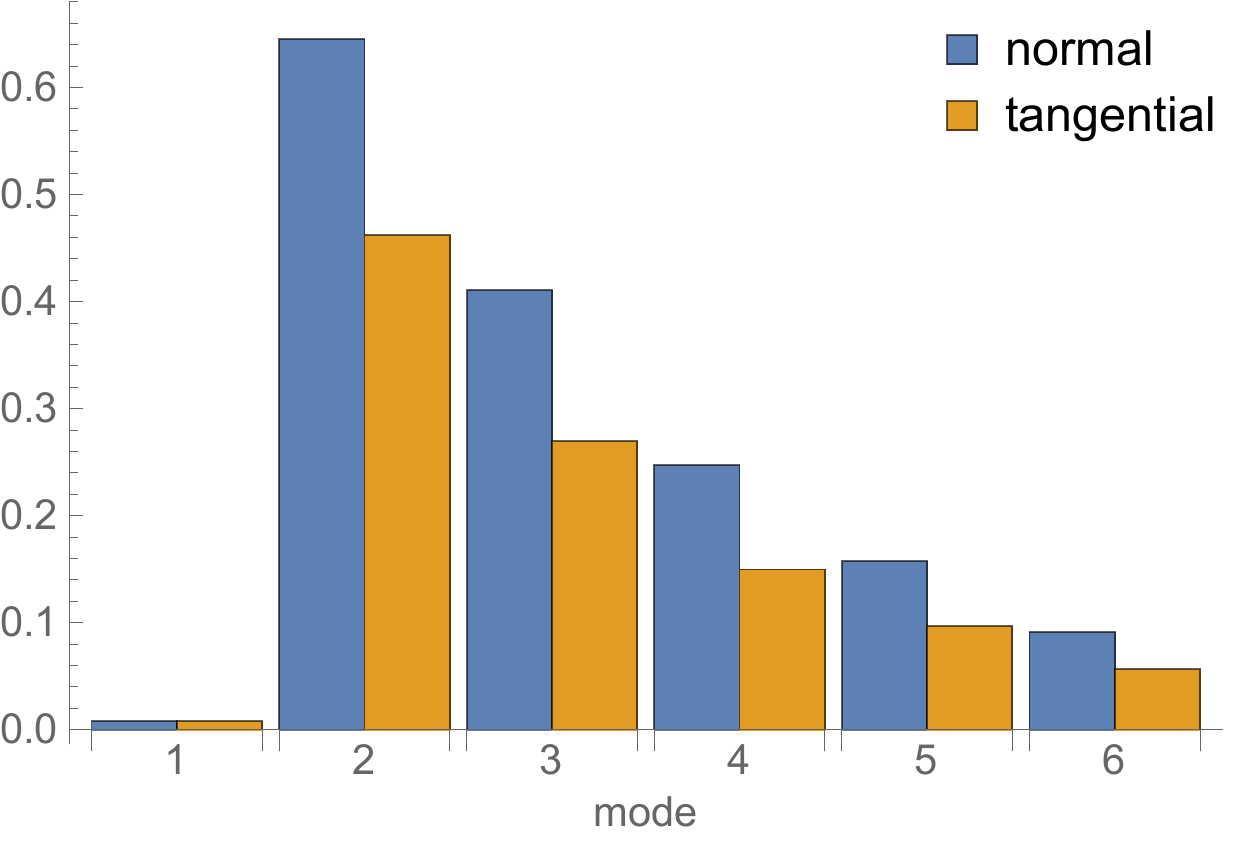}\\
(c) & (d)\\
\end{tabular}
\caption{Flow and stresses for the scenario shown in \fig{vessel geometry} and with parameters of \tbl{example scenario}.
(a,b) Fluid flow near the robot, indicated by the gray disk, in the section of the vessel indicated by the dashed rectangle in \fig{vessel geometry}b.  The green lines indicate the robot's orientation. Arrows show streamlines of the flow and colors show the flow speed. Positions on the axes are in microns. (a) Fluid velocity with respect to the vessel. Velocity is zero at the vessel wall, and matches the motion of the robot at its surface. (b) Fluid velocity with respect to the robot. Velocity is zero at the robot surface.
(c) Stress vectors on the robot surface. The arrow lengths show the magnitude of the stresses, ranging from $0.8\,\pascal$ to $3.9\,\pascal$, at the locations of sensors on the robot surface. The green line indicates the robot's orientation. 
(d) Relative magnitudes of Fourier coefficients of the first six modes of normal and tangential components of stress on the robot's surface, from \eq{normalized Fourier}.
}\figlabel{flow}\figlabel{Fourier}
\end{figure}

\sectlabel{samples}

We generalize this example by creating a set of samples with vessel diameters and flow speeds similar to those of small blood vessels, as indicated in \tbl{fluid parameters}.  
These samples are variations of the situation shown in \fig{vessel geometry} with parameters chosen uniformly at random according to:
\begin{itemize}

\item Maximum inlet fluid speed between 200 and $1000\,\micron/\second$, equally likely flowing to the right or left.

\item Vessel diameter between 5 and $10\,\micron$.

\item Vessel segment length between 18 and $20\,\micron$.

\item Position of robot's center:
\begin{itemize}
\item horizontally, within $2\,\micron$ of the middle of the vessel's length, 
\item vertically, between the center of the vessel and a minimum distance of $0.5\,\micron$ between the robot surface and vessel wall.
\end{itemize}

\item Robot orientation between 0 to 360 degrees.
\end{itemize}
We created 1000 such samples. Of these, 800 samples were used to train the models described below, and the remaining 200 samples were used to test the models.

\section{Estimating Robot and Vessel Geometry}
\sectlabel{geometry}

This section describes stress-based estimates of robot orientation and position in a vessel, and the vessel's diameter. 
Together, these estimates give the robot's relation to the vessel with modest computational cost, as described in \sectA{computation}.

\subsection{Robot Orientation}
\sectlabel{orientation}

Objects rotate as fluid pushes them through a vessel. 
As seen in \fig{Fourier}c, 
the robot has large tangential stress on its side closest to the vessel wall. 
This stress arises from the large velocity gradient between the robot's moving surface and the stationary fluid at the wall, as illustrated in \fig{flow}a. Thus the location on the robot's surface with largest tangential stress magnitude identifies the direction to the nearest vessel wall.

The simplest orientation estimate based on this observation is to use the location of the robot's stress sensor with largest tangential stress magnitude. However, this approach is limited in resolution to the spacing between sensors, and is sensitive to noise in a single sensor. A more robust implementation uses the angle, \angleTExtreme, with respect to the front of the robot that maximizes the magnitude of an interpolation of tangential stress measurements around the surface. Specifically, we use the interpolation based on low-frequency Fourier modes given by \eq{interpolation}. This interpolation combines information from multiple sensors to reduce noise, and can identify locations that lie between sensor positions for improved resolution. The estimated direction to the wall, $\angleToWall$, is taken to be the direction of the  normal vector at the location  \angleTExtreme\ on the surface. For circular robots the surface normal vector aligns with the center of the robot so $\angleTExtreme=\angleToWall$. However, these angles can differ for other shapes, as discussed in \sectA{elongated}.

As shown in \fig{vessel geometry}, the robot's motion is nearly parallel to the vessel wall. Thus, the robot can estimate its direction of motion as perpendicular to its estimated direction to the wall. More specifically, \fig{Fourier}c shows that, near the direction to the wall, the normal stress is negative in the direction of motion. This observation allows the robot to determine whether the direction of motion is $+90^\circ$ or $-90^\circ$ from the direction to the wall, with details given in \sectA{direction of motion}.

We evaluated the interpolation method on the samples described in \sect{samples}. The mean magnitude of the errors in direction to the wall and direction of motion were both less than $1^\circ$. 
The largest errors occur for robots near the center of the vessel, where the tangential stresses on the sides facing the opposite walls are nearly the same. 
However, for robots near the center, the wall in either direction is at about the same distance from the robot, so such errors have little significance.

\subsection{Relative Position}
\sectlabel{relative position}

The pattern of surface stresses are significantly different for robots near the center of the vessel and those near its wall. 
A useful way to express this variation independently of vessel diameter is through the robot's \emph{relative position} in the vessel, defined as
\begin{equation}\eqlabel{relative position}
\relPos = \frac{|y_c|}{d/2-r} 
\end{equation}
where $y_c$ is the position of the robot's center relative to the central axis of the vessel, as shown in \fig{vessel geometry}b. Relative position ranges from 0, for a robot at the center of the vessel, to 1, for a robot just touching the vessel wall. 
We estimate relative position from Fourier coefficients of the stress measurements by training a regression model (see \sectA{regressions}).

\begin{figure}
\centering \includegraphics[width=\figwidth]{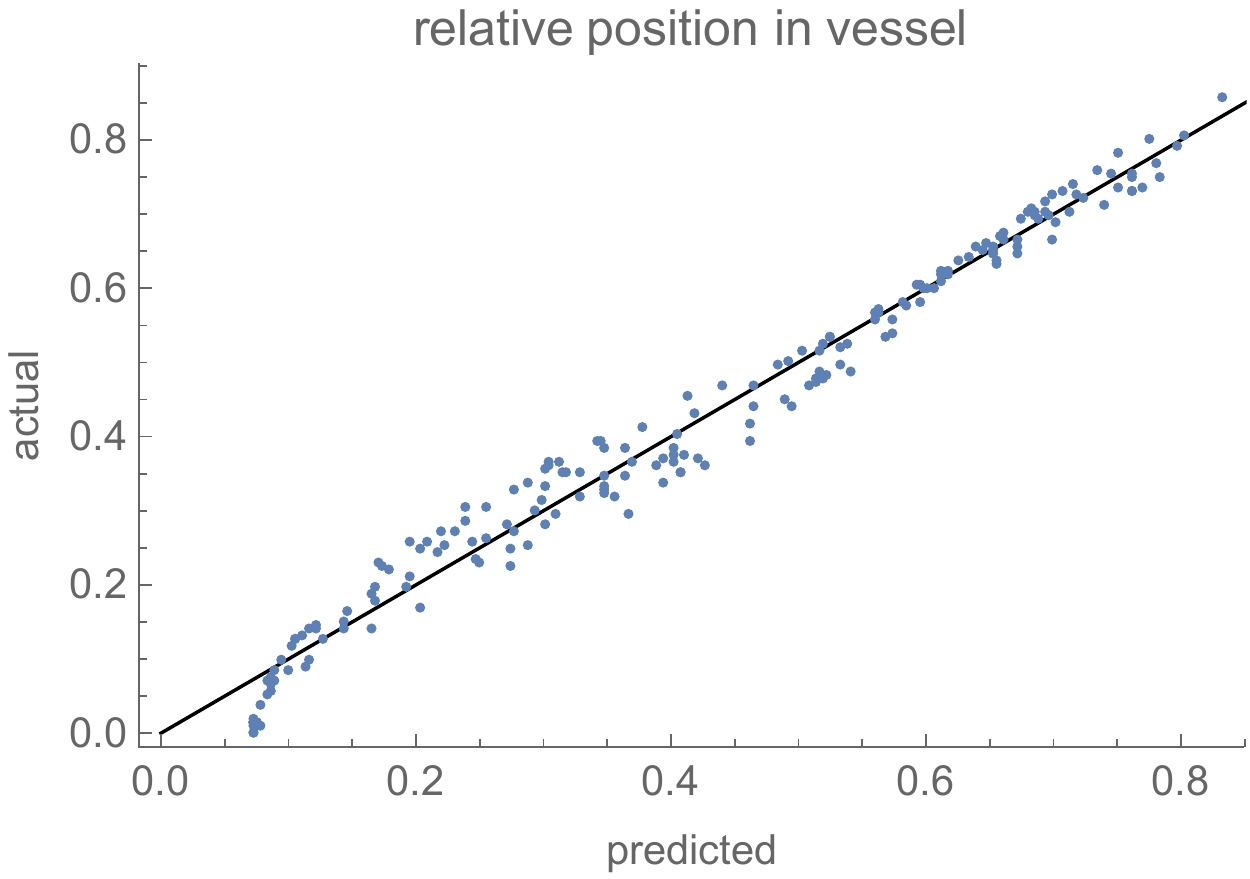}
\caption{Predicted and actual relative positions in the vessel. The straight line indicates perfect prediction, i.e., where predicted and actual values are the same.}\figlabel{relative position}
\end{figure}

\fig{relative position} compares predicted and actual values for the test samples described in \sect{samples}. The root mean square (\RMS) error for the test samples is $0.032$. Thus by measuring surface stress, a robot can fairly accurately determine its relative position. 
The regression is least accurate for robots close to the center of the vessel, i.e., with relative position less than about $0.1$.

\subsection{Vessel Diameter}
\sectlabel{diameter}

For flow at a given speed, wider vessels have smaller fluid velocity gradients than narrower ones. This leads to different stress patterns for robots at given relative positions in those vessels, especially for robots close to the wall, where gradients are largest. A robot can exploit these differences with a regression relating vessel diameter to Fourier coefficients of the stress, as described in \sectA{regressions}.

\begin{figure}
\centering \includegraphics[width=\figwidth]{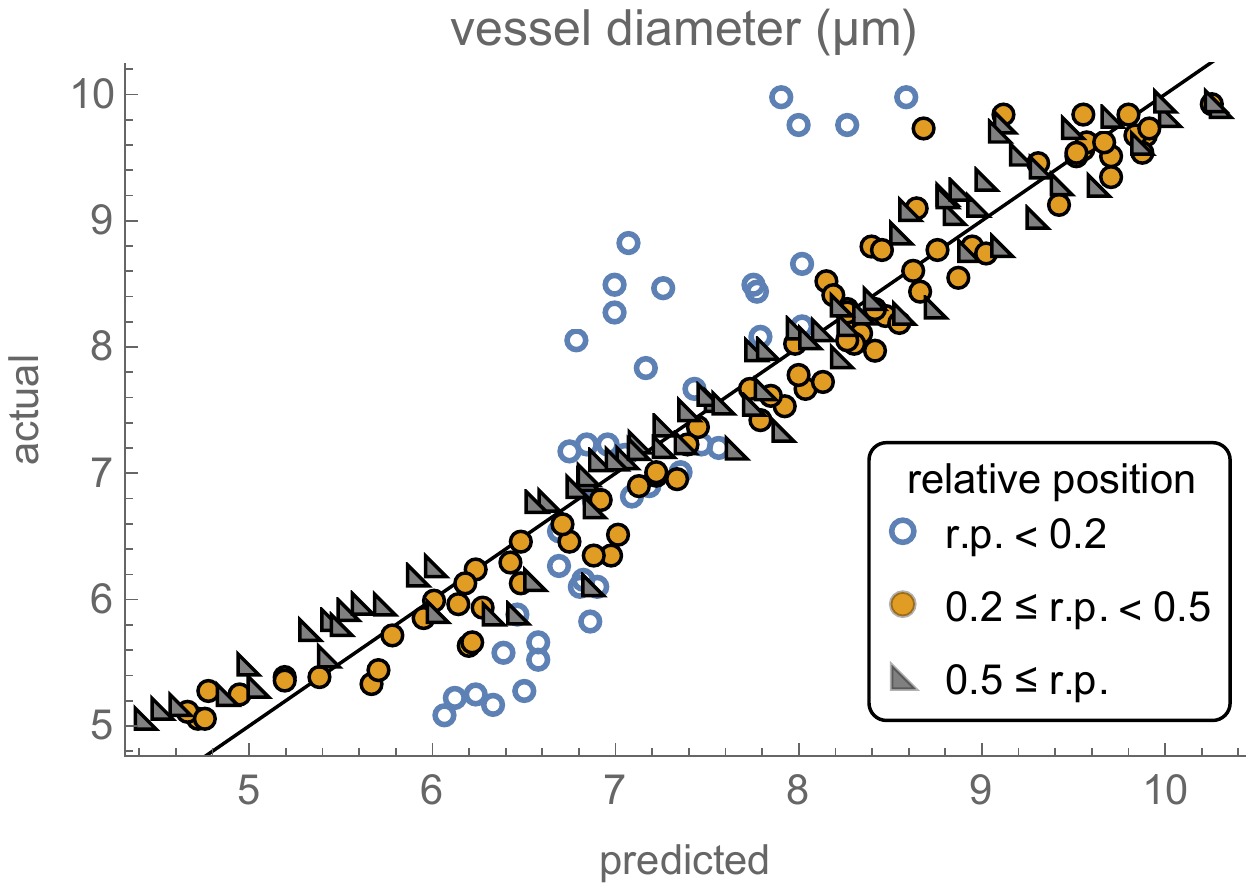}
\caption{Predicted and actual vessel diameters. The straight line indicates perfect prediction, i.e., where predicted and actual values are the same. The point markers group the instances by their relative positions, as described in the legend.}\figlabel{vessel diameter}
\end{figure}

\fig{vessel diameter} compares predicted and actual values for the test samples described in \sect{samples}. The root mean square (\RMS) error for the test samples is $0.51\,\micron$, within about 10\% of the actual diameters considered here.  
The largest errors are for robots near the center of the vessel.
For example, the \RMS\ errors are $0.9\,\micron$ and $0.3\,\micron$ for the 41 and 159 test samples with relative positions smaller than and greater than or equal to $0.2$, respectively.

\subsection{Distance to Vessel Wall}

\begin{figure}
\centering \includegraphics[width=\figwidth]{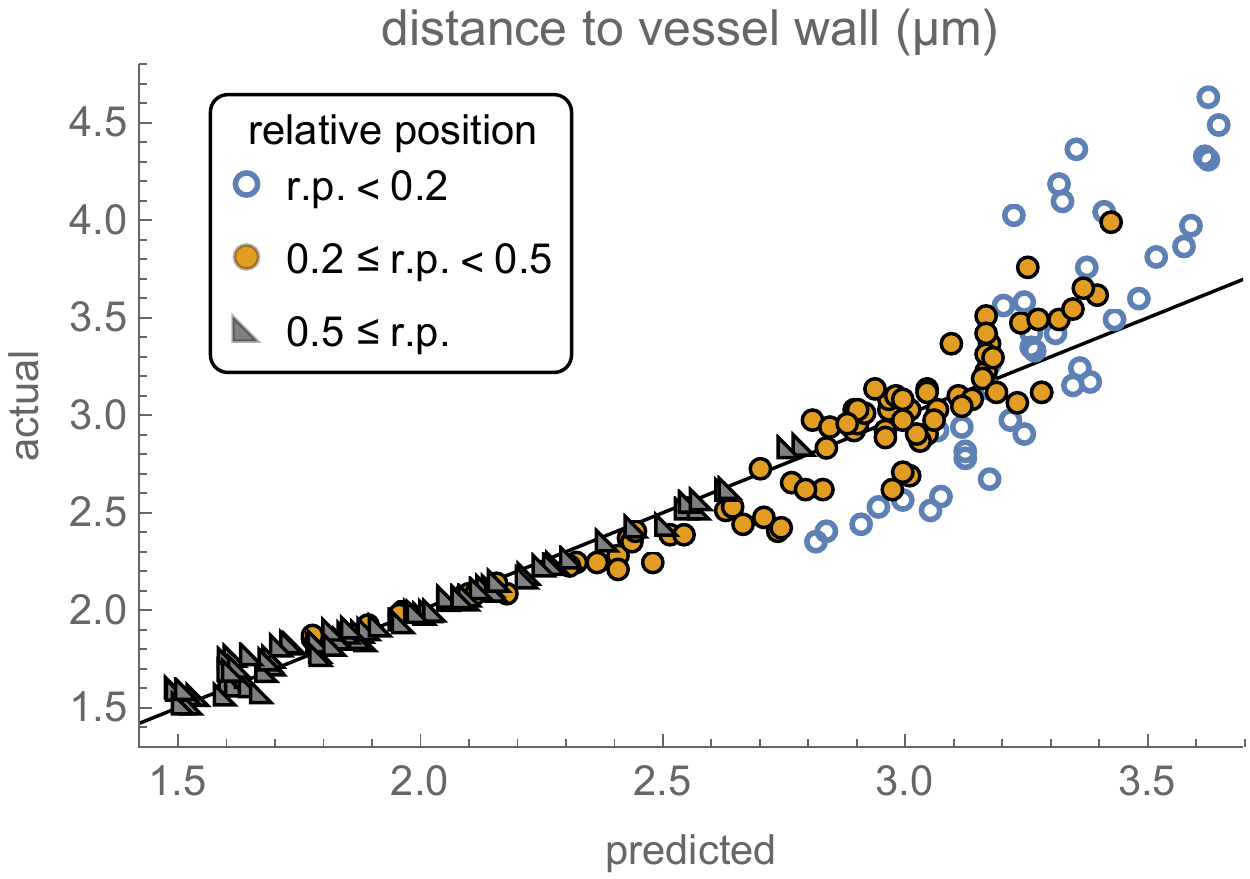}
\caption{Predicted and actual distance between the robot's center and the nearest point on the vessel wall. The straight line indicates perfect prediction, i.e., where predicted and actual values are the same. The point markers group the instances by their relative positions, as described in the legend.}\figlabel{distance to wall}
\end{figure}

\eq{relative position} relates relative position, diameter and the distance of the robot's center to the vessel wall (i.e., $d/2-|y_c|$ from \fig{vessel geometry}b). Thus combining estimates of relative position and vessel diameter gives an estimate of the robot's distance to the wall.

\fig{distance to wall} evaluates this procedure for the test samples of \sect{samples}. Estimates are accurate near the wall, allowing the robot to identify when it is close to the wall. On the other hand, the prediction error is particularly large for robots near the center of the vessel. In particular, the \RMS\  errors are $0.5\,\micron$, $0.2\,\micron$ and $0.05\,\micron$ for the 41, 81 and 78 test samples with relative positions less than $0.2$, between $0.2$ and $0.5$ and at least $0.5$, respectively.

\section{Estimating Robot Motion}
\sectlabel{motion}

This section discusses how the robot can estimate its motion in the vessel.
For Stokes flow, the stresses and hence the Fourier coefficients are linear functions of the robot's speed and angular velocity. The stresses have a more complex dependence on the geometry, i.e., vessel diameter and robot position within the vessel. One approach to estimating motion would be to train regression models of these relationships, similar to the procedure for estimating position. This section describes an alternate, and simpler, approach: estimating angular velocity from how stresses change with time, and combining that estimate with a regression model to estimate the robot's speed.

\subsection{Angular Velocity}
\sectlabel{angular velocity}

For straight vessels, the rate of change in the direction to the vessel wall equals the robot's angular velocity. 
A robot's motion in the vessel is mainly directed downstream, with relatively slow change in its distance to the wall. This means the pattern of stress on its surface, measured from the front of the robot, maintains roughly the same shape but shifts around the robot due to its rotation.
Thus the robot can estimate its angular velocity from the angle $\Delta \theta$ around its surface that the stress pattern moves in a short time interval $\Delta t$: $\omegaRobot \approx -\Delta \theta/\Delta t$. 
A robot can determine $\Delta \theta$ from the correlation between the surface stresses at these two times, as described in \sectA{estimating angular velocity}.

We evaluate this method with 100 of the samples described in \sect{samples}. The motion of each sample was evaluated for $\Delta t = 5\,\millisecond$. In each case we determine the angle $\Delta \theta$ maximizing the correlation between the stresses before and after the motion.
The maximal correlations were above $99.9\%$ for all the samples, indicating negligible change in the shape of the stress pattern during this time interval.
\fig{robot angular velocity} compares predicted and actual values of the angular velocity for these samples. The mean error for the estimates is $0.51\,\radian/\second$.

\begin{figure}
\centering \includegraphics[width=\figwidth]{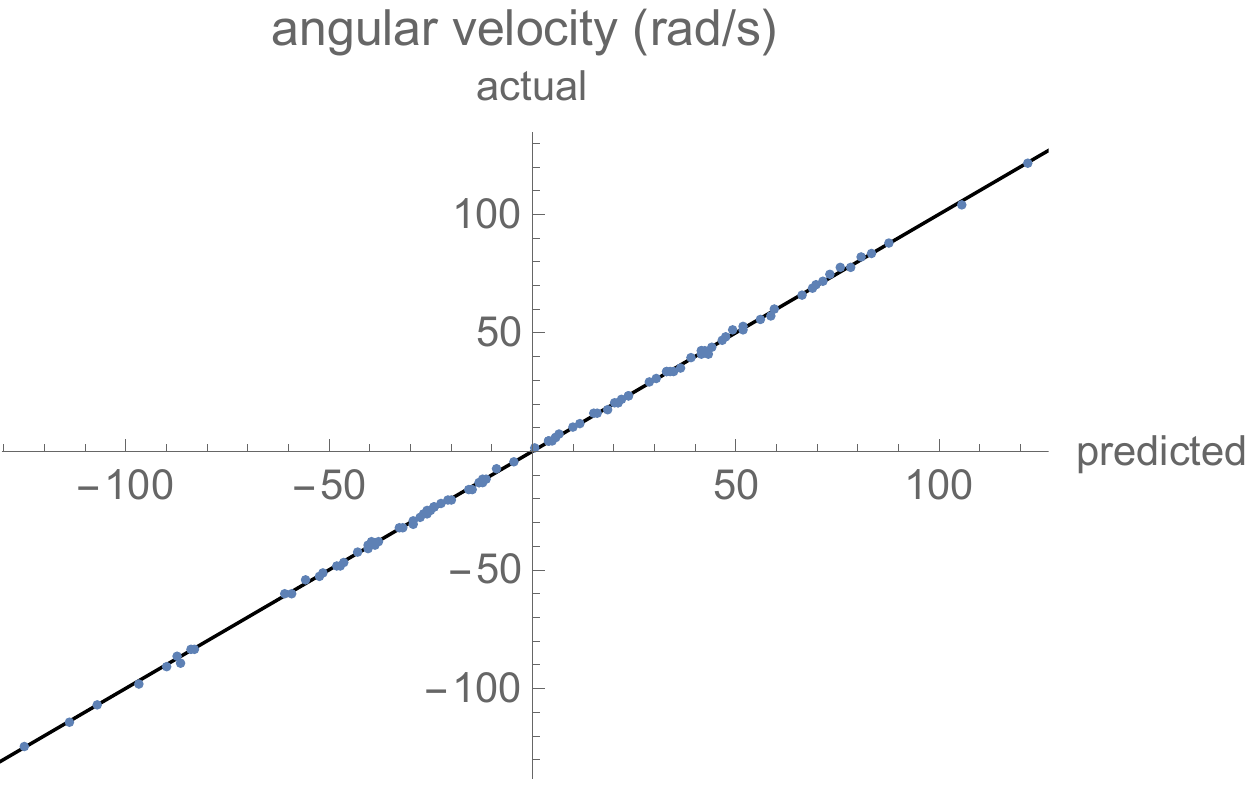}
\caption{Predicted and actual robot angular velocities. The predictions are from the average of shifts in the patterns of normal and tangential stresses over $5\,\millisecond$. The straight line indicates perfect prediction, i.e., where predicted and actual values are the same.}\figlabel{robot angular velocity}
\end{figure}

\subsection{Speed Ratio}

The flow in a vessel has a parabolic speed profile~\cite{happel83}. Thus, near the center the flow is relatively fast and the gradient of speed with distance from the vessel wall is relatively small. Conversely, near the vessel wall the speed is low and gradient large. These variations mean a robot near the center moves relatively quickly but rotates slowly, since the flow on either side is fairly similar. On the other hand, a robot near the wall moves slowly, but the large gradient makes it rotate rapidly. 
Thus the ratio of speed through the fluid to rotational velocity at the robot's surface
\begin{equation}\eqlabel{speed ratio}
\speedRatio = \left| \frac{\vRobot}{\omegaRobot r} \right|
\end{equation}
is large for a robot near the center of the vessel and small for a robot near the wall. Since both robot speed and angular velocity are proportional to overall fluid speed and viscosity, their ratio is independent of these factors. 

We find that the speed ratio is close to a linear function of the odds ratio, $(1-\relPos)/\relPos$, of the relative position defined by \eq{relative position}. 
Thus a useful model for estimation is a least-squares fit to the training samples, described in \sect{samples}, which gives
\begin{equation}\eqlabel{speed ratio fit}
\speedRatio = a + b \left( \frac{1-\relPos}{\relPos} \right)
\end{equation}
with $a=3.1 \pm 0.2$ and $b=5.41\pm0.03$, where the ranges indicate the standard error of the values.

For the test samples, this fit gives median relative error of $8\%$ for the speed ratio. The largest errors arise when the robot is very close to the center of the vessel (i.e., cases with $\relPos < 0.01$).

\subsection{Speed}
\sectlabel{speed}

\begin{figure}
\centering \includegraphics[width=\figwidth]{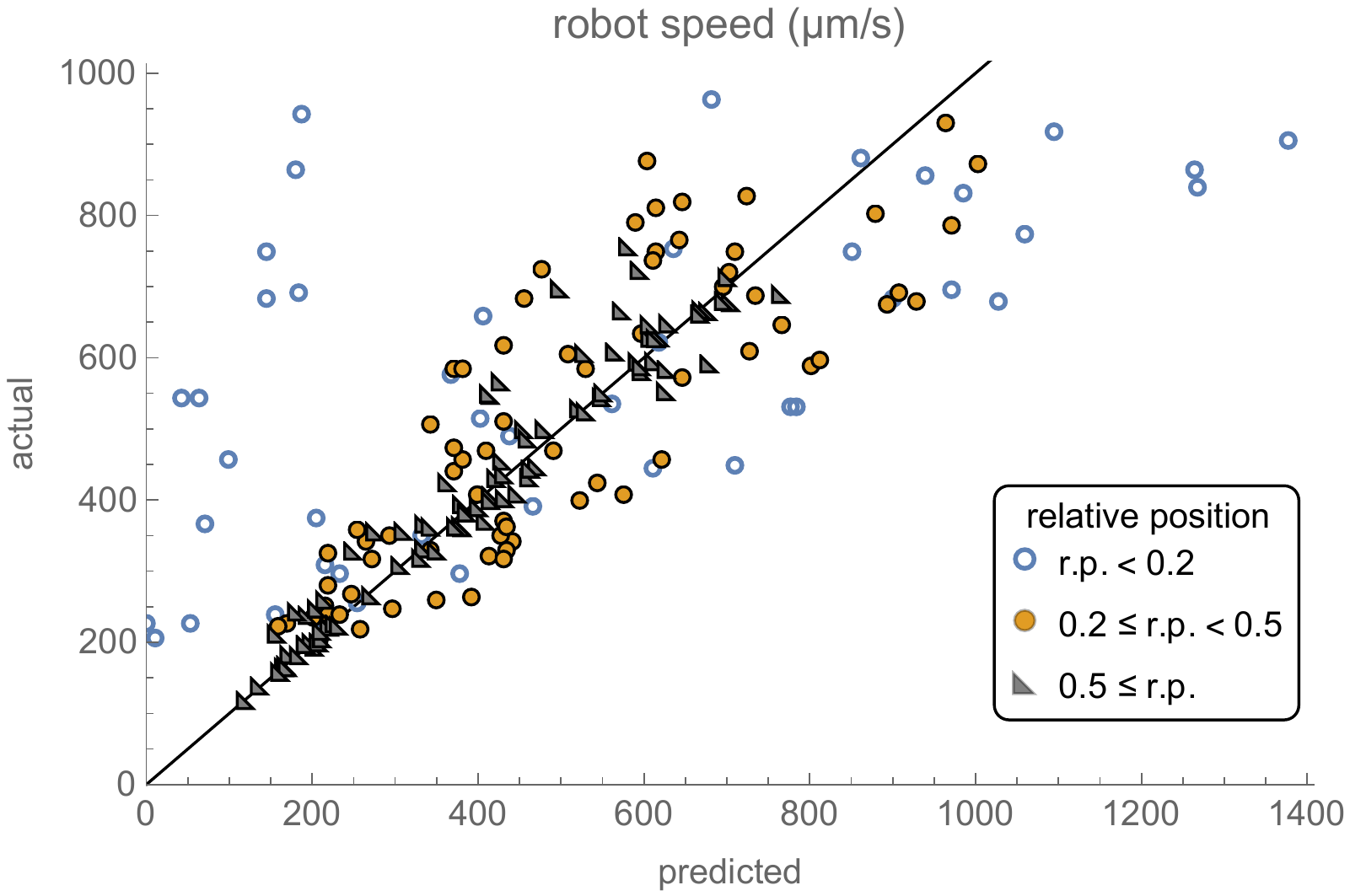}
\caption{Predicted and actual robot speeds. The straight line indicates perfect prediction, i.e., where predicted and actual values are the same. The point markers group the instances by their relative positions, as described in the legend.}\figlabel{robot speed}
\end{figure}

Combining estimates of angular velocity and speed ratio gives an estimate for the robot's speed through the vessel via \eq{speed ratio}. 
\fig{robot speed} compares predicted and actual speeds for the test instances described in \sect{samples}.
The mean relative error is $20\%$.
The prediction error varies significantly with robot position. Specifically, mean relative errors are $43\%$, $21\%$ and $7\,\%$ for the 41, 81 and 78 test samples with relative positions less than $0.2$, between $0.2$ and $0.5$ and at least $0.5$, respectively.
The large error for small relative positions, i.e., when the robot is close to the center of the vessel and has a relatively small angular velocity, is due to unreliable estimates of the speed ratio in such cases.

\section{Discussion}

This paper shows how microscopic robots can use surface stresses to estimate their direction and distance to the nearest vessel wall, the vessel diameter, and how fast they are moving. 
Stress-based navigation is well-suited to micron-size robots in capillaries. This is because such robots are both small enough to fit in the vessels and large enough to have useful signal-to-noise when averaging over the time a robot moves a distance comparable to its size, as discussed in \sectA{noise}. Neither larger robots nor significantly smaller devices, such as nanoparticles, satisfy both conditions.

Measuring surface stresses could aid a variety of applications.
For example, decreasing estimates of vessel diameter indicate a narrowing vessel where the robot may become stuck. 
By responding to this information, the robot could avoid damage to itself or its surroundings. 

An orientation estimate allows a robot to determine its upstream and downstream sides, which could improve chemical gradient measurements~\cite{dusenbery09}.
For robots with locomotion capability, knowing its orientation allows the robot to move directly toward the vessel wall, thereby reaching the wall more rapidly and with less energy use than a blind search. 
Furthermore, orientation estimates could help control robot locomotion, e.g., to maintain a fixed orientation with respect to the vessel in spite of Brownian motion and fluid torques. This could simplify a subsumption architecture~\cite{brooks91a} for microscopic robots, since higher-level controls, operating at longer time scales, would then not need to account for these unintended changes in orientation.

Extensions to the methods described here may improve accuracy, especially for robots near the center of the vessel where the estimates are least accurate. 
For example, the regressions discussed here use the relative magnitudes of the Fourier coefficients, which do not depend on robot orientation or the fluid's speed and viscosity.
However, this normalization also discards any information on vessel geometry in the phases and magnitudes of the Fourier coefficients.
Developing estimates using this information may improve accuracy, but also will require more complex processing or stronger assumptions about the environment than needed for estimates based on relative magnitudes. 
For instance, the orientation estimate of \sect{orientation} could adjust phases to a standard reference frame, and assumptions of slowly changing fluid speed and viscosity would allow attributing abrupt changes in magnitude to a change in the robot's position or vessel diameter.

This paper used a few Fourier modes to represent surface stresses.
An open question is whether other representations can extract information more effectively. 
For example, a robot near the wall has large stresses over a small area close to the wall. Resolving those fine spatial details requires many Fourier modes, so using a few wavelets at coarse and fine scales may be more effective.
Moreover, if there is additional useful information in the full 3D stress pattern (see \fig{3D stresses}), regressions based on spherical Fourier expansions~\cite{orszag74} or spherical harmonics~\cite{mohlenkamp99} may improve the estimates.

The robots considered in this paper move in straight vessels containing no other objects. More complex geometries include curved or branching vessels, and multiple objects in the flow. 
While the  methods of this paper give rough estimates in other geometries (see \sectA{generalization}), 
estimators trained on these geometries would likely be more accurate.
In addition, distinguishing among stresses of different geometries may allow robots to identify, e.g., when they reach a branch in the vessel.

The robots considered here are rigid. Fluid stresses are far smaller than the elastic modulus of materials microscopic robots could use. Thus the assumption of a fixed shape is reasonable.
However, some robot can actively alter their shapes~\cite{castano00}. For such robots, surface forces could indicate when and how to do so, e.g., to assume a more elongated shape~\cite{hogg16} when the robot enters a narrow vessel.

For determining the robot's position in the vessel, the methods in this paper use the stress on the robot at a given time. Monitoring how stresses change as the robot moves may improve the estimates. 
For example, stresses increase in magnitude as a robot approaches the vessel wall. 
Moreover, robots attached to the vessel wall could use changing stresses to detect objects passing in the fluid. More generally, these estimation methods could interpret stress measurements from sensors in the walls of engineered microfluidic vessels.
Additional changes in stress arise for elongated or shape-changing robots.
For example, the pattern of stress on elongated robots changes as the fluid rotates them while they move. Moreover, shape-changing robots experience different stresses as they change shape, which they could use as active probes of nearby geometry.

This paper considered simple fluids in small vessels with rigid boundaries.
The viscosity of some biological fluids varies significantly with the size and surface coatings of objects in those fluids~\cite{lai09}.
Robots may be able to exploit these effects by, for example, using stress to detect how their motion affects nearby non-rigid boundaries~\cite{trouilloud08} or tuning their size, shape and surface properties to minimize viscosity as proposed for nanoparticle-based drug delivery~\cite{lai09}.

In summary, stresses can provide microscopic robots with useful information about their environments by exploiting the properties of viscous flow. 
More generally, theoretical studies, such as presented in this paper, can quantify likely capabilities of these robots prior to their manufacture, and hence inform their design and help identify suitable applications for them.

\section*{Acknowledgements}

I thank Robert A. Freitas~Jr., Ralph C. Merkle, Matthew S. Moses, Daniel Reda and James Ryley for helpful discussions.

\newpage
\appendix

\numberwithin{equation}{section}
\numberwithin{figure}{section}
\numberwithin{table}{section}

\begin{center}
{\huge \textbf{Appendices}}
\end{center}

\section{Fluid and Robot Motion}
\sectlabel{flow}

This appendix describes the nature of the flow and the forces acting on the robot, and simplifying approximations for numerically computing these forces over millisecond time scales.

\subsection{Fluid Flow}

For fluids moving slowly in small vessels, viscous forces are significantly larger than inertial forces. This leads to smooth, laminar motion called Stokes flow~\cite{kim05,happel83}.
Quantitatively, Stokes flow occurs when the flow's Reynolds number
\begin{equation}
\Rey=\frac{u d}{\viscosityKinematic}
\end{equation}
is small. In this expression, $\viscosityKinematic$ is the fluid's kinematic viscosity, $d$ the vessel diameter and $u$ the flow speed. \tbl{fluid parameters} shows $\Rey\ll 1$ for the flows considered here.

For a robot moving slowly in such flows, the fluid motion at each time is close to the static flow associated with its instantaneous position, velocity and orientation~\cite{happel83,kim05}. This quasi-static approximation applies for small values of the Womersley number
\begin{equation}
\Wom=\frac{r}{\sqrt{\viscosityKinematic t}}
\end{equation}
where $r$ is the robot's radius and $t$ is the time over which the geometry changes.
Changes in robot geometry arise from changes in its position or orientation. The time for significant change in the robot's position is of order $r/\vRobot$, which is larger than $r/u$ because a robot moving passively in the fluid moves more slowly than the fastest fluid speed in the vessel. Similarly, significant changes in orientation take time of order $1/\omegaRobot$, which is also larger than $r/u$. Thus a lower bound on the time for significant geometry change is $t>r/u$.
For the robot motion considered here, with the parameters of \tbl{robot parameters}, $\Wom < 0.05$ so quasi-static Stokes flow is a good approximation.

Evaluating the flow requires specifying boundary conditions at the ends of the vessel segment and on the wall. Specifically, we take the incoming flow to be full-developed Poiseuille flow~\cite{happel83}. This gives a parabolic velocity profile at the inlet, with maximum speed $u$ at the center of the vessel. A pressure gradient drives the flow, so shifting the pressure by a constant leaves the flow unchanged. For definiteness, we set the pressure to zero at the outlet.
The vessel walls are no-slip boundaries, where the fluid velocity is zero. The flow velocity matches the robot's motion at each point on the robot's surface, expressed in terms of the robot's center-of-mass velocity and angular velocity.

\subsection{Robot Motion}

A robot is subject to forces from the fluid~\cite{happel83}, Brownian motion and gravity. 
For simplicity, we consider the robots to be neutrally-buoyant so gravity does not affect the motion. 

To estimate the importance of Brownian motion, the diffusion coefficient of a sphere is~\cite{berg93}
\begin{equation}
D = \frac{\BoltzmannConstant \Tfluid}{6\pi \viscosity r} \approx 10^{-13}\,\meter^2/\second
\end{equation}
with the parameters of \tbl{fluid parameters}, and where \BoltzmannConstant\ is Boltzmann's constant.
The Peclet number
\begin{equation}
\Pec = \frac{\vRobot r}{D}
\end{equation}
characterizes the relative importance of convective and diffusive motion over distances comparable to the robot size.
At the lower range of velocities considered here, $\vRobot \approx 100\,\micron/\second$, $\Pec = 1000$ which is large compared to 1. Thus diffusion is not important.

The rotational diffusion coefficient of a sphere is~\cite{berg93}
\begin{equation}
D_\text{rot} = \frac{\BoltzmannConstant \Tfluid}{8\pi \viscosity r^3} \approx 0.1\,\radian^2/\second
\end{equation}
For the parameters of \tbl{robot parameters},  $\omegaRobot \approx 100\,\radian/\second$ so for a change in orientation of $\theta=1\,\radian$, $\omegaRobot \theta/D_\text{rot} \approx 1000$, showing rotational diffusion is not significant.

Since diffusive motion is relatively small over the times considered here, this paper treats motion as deterministically arising from the fluid forces on the robot. For the time scales considered here, viscous forces keep the robot close to its terminal velocity in the fluid~\cite{purcell77}. Thus we determine robot's velocity and angular velocity as the values giving zero net force and torque on the robot.

\subsection{Two and Three Dimensional Flow}

\begin{figure}
\centering
\begin{tabular}{cc}
\includegraphics[width=\mfigwidth]{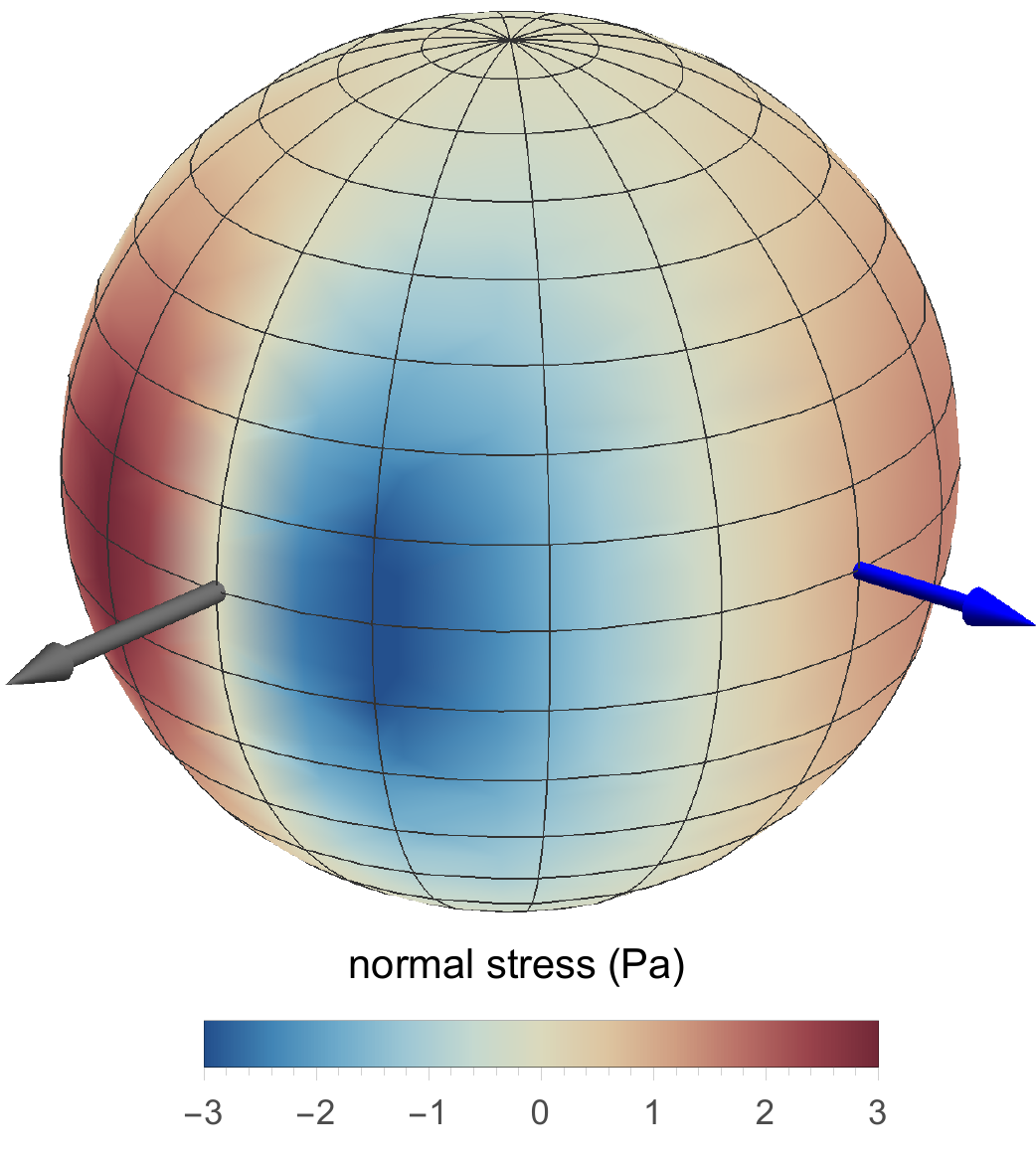} & \includegraphics[width=\mfigwidth]{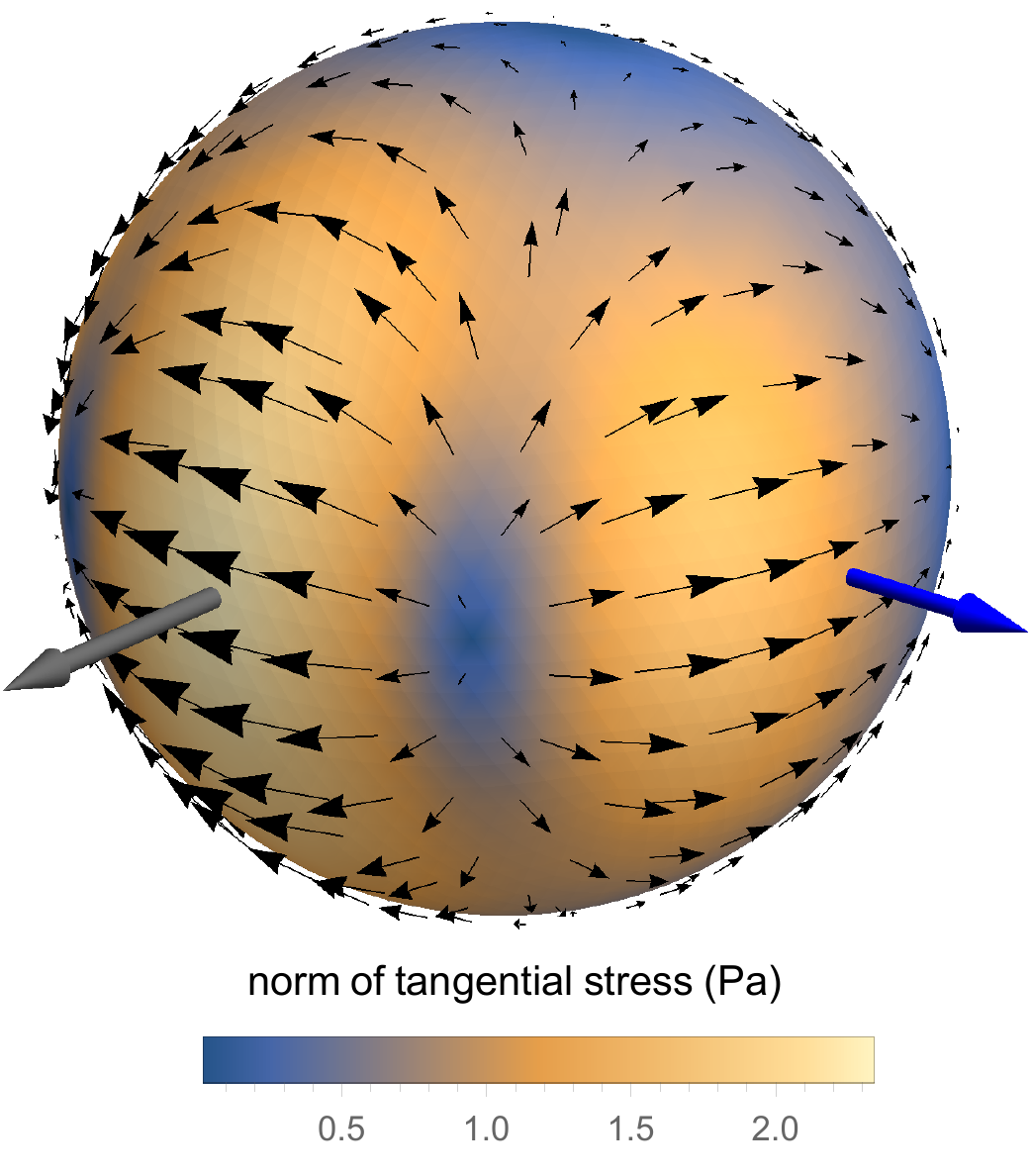}\\
(a) & (b) \\
\end{tabular}
\caption{Stresses on the sphere in a vessel with the geometry and viewpoint of \fig{vessel geometry}a and parameters of \tbl{example scenario}.
The blue arrows, on the right of each sphere, indicate the direction of motion, as also shown in \fig{vessel geometry}. The gray arrows indicate the direction to the nearest point on the vessel wall. 
(a) Normal stress: positive and negative values correspond to tension and compression, respectively. (b) Tangential stress: arrows show the direction and relative size of the tangential stress vector and the color indicates its norm.}\figlabel{3D stresses}
\end{figure}

\fig{3D stresses} shows the pattern of stress on the robot's surface for the scenario of \tbl{example scenario}.
The figure decomposes the stress into components normal and tangential to the robot's surface. 
The fluid flow is driven by a pressure gradient in the vessel. Adding an arbitrary constant to the pressure does not change the flow but contributes to the normal force on the robot surface. To remove this arbitrariness, the robot select's this constant so the total measured normal stress on its surface is zero.

The stress is relatively large on the side of sphere facing the nearby vessel wall. In addition, the largest variation of the stress is in the azimuthal direction around the sphere's equator. In particular, the largest magnitude of the tangential stress is at the point of the sphere closest to the vessel wall. The sphere's direction of motion is nearly $90^\circ$ around the equator from that closest point.

\begin{figure}
\centering 
\begin{tabular}{cc}
\includegraphics[width=\mfigwidth]{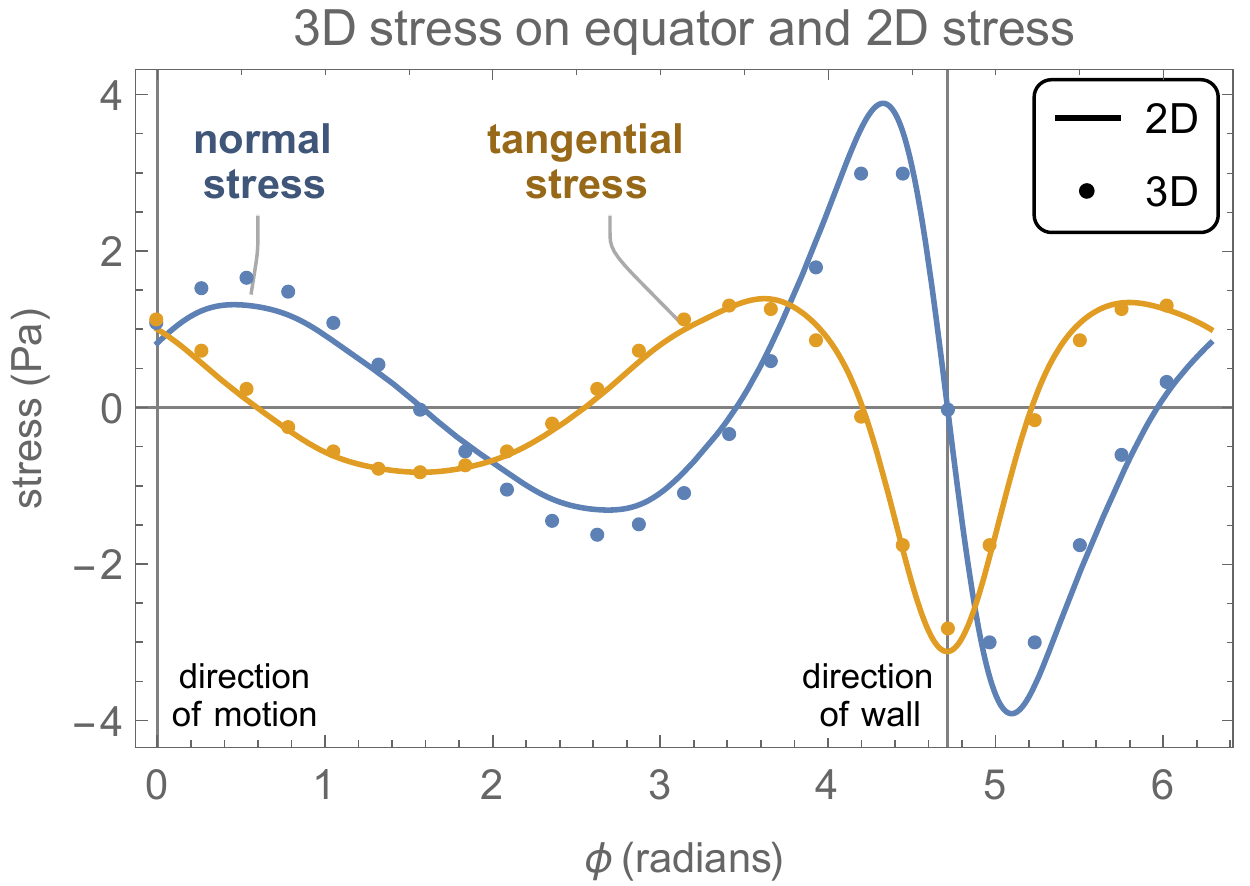} &
 \includegraphics[width=\mfigwidth]{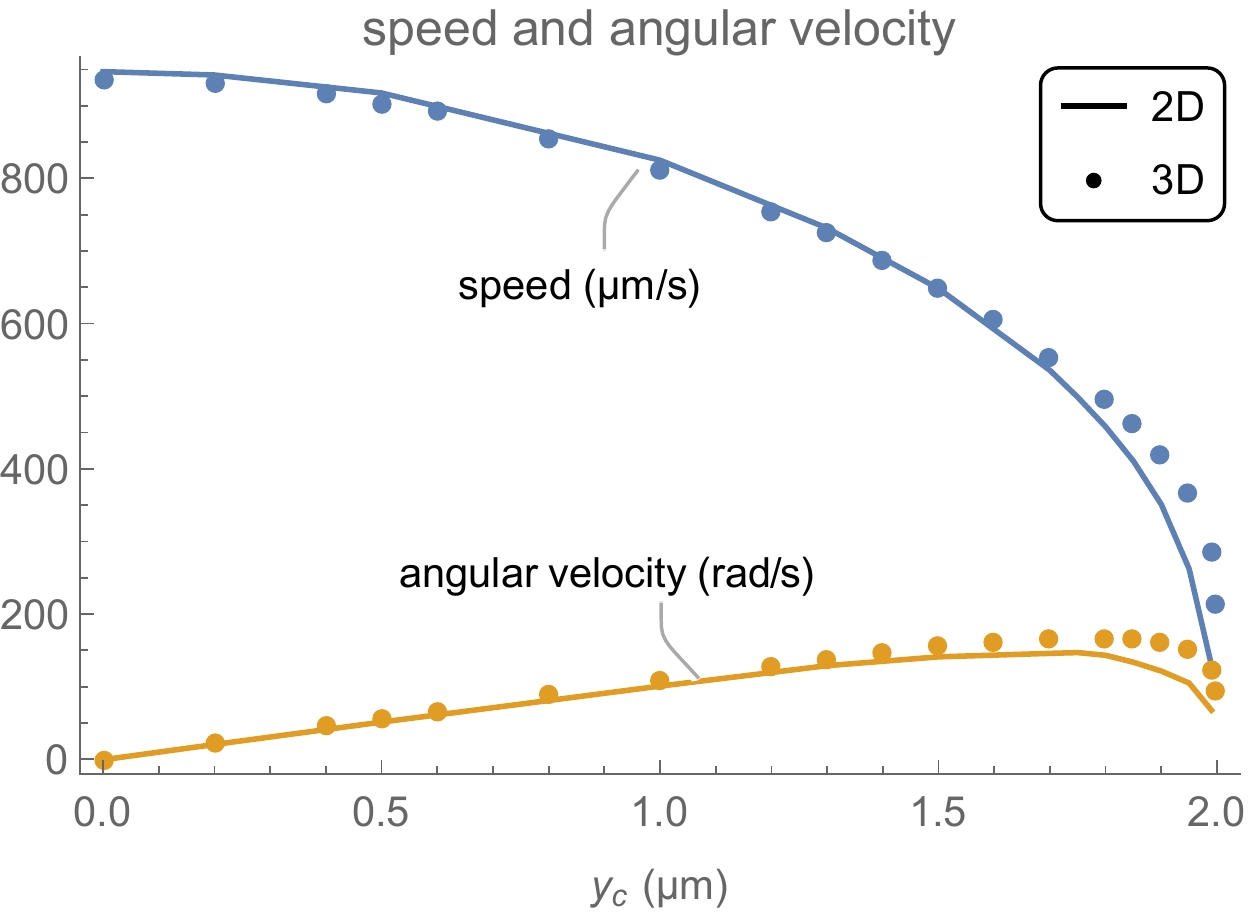}\\
 (a) & (b)\\
\end{tabular}
\caption{Comparison of two and three-dimensional behavior. (a) Normal and tangential stresses for the scenario of \fig{3D stresses}, as a function of azimuthal angle $\phi$ from $0$ to $2\pi$, measured from the direction of the cylinder axis, which is nearly parallel to the robot's velocity. The two vertical lines indicate the directions of motion and the nearest point on the vessel wall. Positive tangential stress is in the direction of increasing $\phi$, i.e., counterclockwise.
The stress components correspond to the stress vectors shown in \fig{Fourier}c.
(b) Speeds and angular velocity for the scenario of \tbl{example scenario} but varying the position, $y_c$, of the sphere's center within the vessel, ranging from zero, with the sphere in the center of the vessel, to slightly less than $d/2-r=2\,\micron$, with the sphere almost touching the vessel wall.
}\figlabel{3D and 2D}
\end{figure}

\fig{3D and 2D}a compares the stress from 2D fluid flow with the stress around the sphere's equator of the 3D flow. In the latter case, by symmetry, the tangential stress is directed around the equator, 
thereby matching the direction of the tangential stress of the 2D case. 
As an additional comparison, \fig{3D and 2D}b shows the robot's speed through the vessel, $\vRobot$, and angular velocity, $\omegaRobot$, as a function of position in the vessel, $y_c$, for the geometry of \fig{vessel geometry}. In these cases, the robot's motion is very nearly parallel to the cylinder's axis because the transverse component of the robot's velocity is less than $1\,\micron/\second$.
In both these comparisons, the 2D and 3D calculations give similar results. 
Thus this paper uses numerical solutions of quasi-static Stokes flow in two dimensions.

\section{Estimation From Stress Measurements}
\sectlabel{measurements}

The robot estimates its position and motion within the vessel via their relationship to its stress measurements. We determine this relationship from samples of stress measurements in known geometries. 
If robots could be fabricated, these samples could be obtained experimentally. For example, the forces on microfluidic devices can be measured externally~\cite{wu10}, and \invivo\ imaging~\cite{medina-sanchez17} could record robot behavior in biological environments, thereby calibrating the robot's sensors in realistic situations.
Since this experimental approach is not yet feasible, this paper instead creates samples from numerical solution of the flow with random parameter choices described in \sect{samples}. 

\begin{figure}
\centering 
\begin{tabular}{cc}
\includegraphics[width=\smallfigwidth]{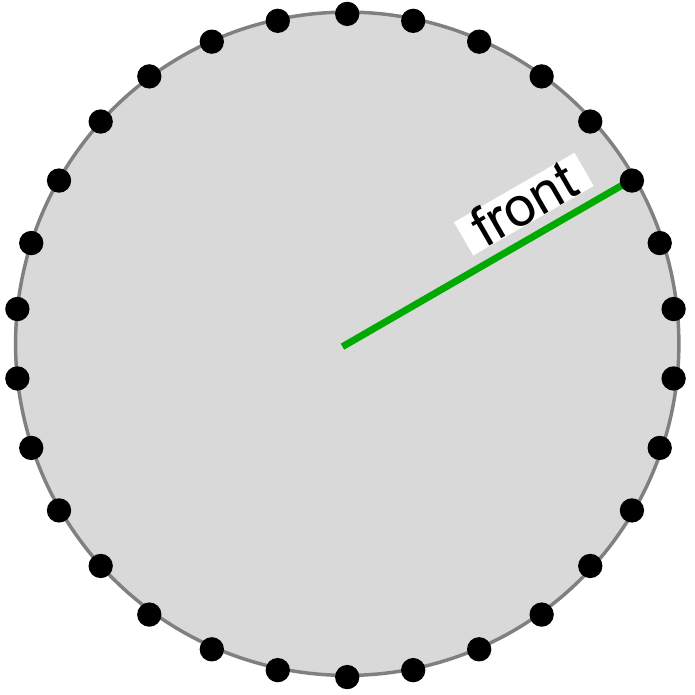} &
 \includegraphics[width=\mfigwidth]{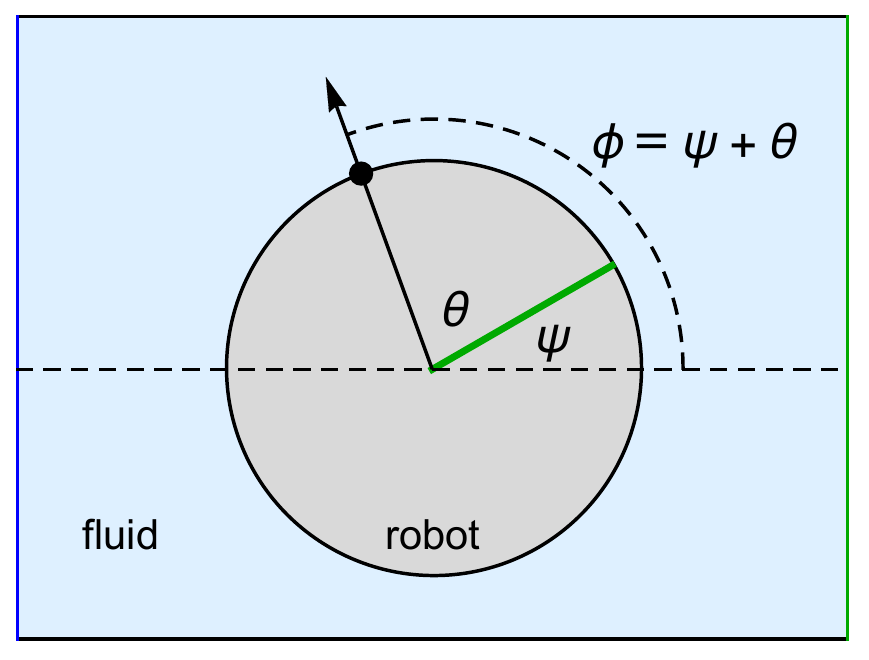}\\
 (a) & (b)\\
\end{tabular}
\caption{Robot angular measurement convention. (a) Robot, indicated as a gray disk, with $n=30$ stress sensors, shown as points, spaced uniformly around its surface. The green line indicates the front of the robot. Sensors are numbered from 0 to $n-1$, counterclockwise around the robot starting from its front. 
(b) Angles specifying a point on the robot's surface, measured with respect to the $x$-axis ($\phi$) and with respect to the robot's front ($\theta$). The angle $\orientation$ is the robot's orientation, i.e., the angle of its front with respect to the $x$-axis.
}\figlabel{robot with sensors}
\end{figure}

We consider a two-dimensional circular robot with $n$ uniformly spaced stress sensors. Sensor $j$, for $j=0,\ldots,n-1$, is located at angle $2\pi (j/n)$ measured counterclockwise from a specified point on the surface, called the robot's ``front'', as illustrated in \fig{robot with sensors}a.
We denote the stress measured by sensor  $j$ as $\vec{f}_j$, with components normal and tangential to the robot surface. 

The stresses depend on the robot's position and motion in the vessel. These are the properties the robot needs to estimate from its stress measurements.
Complicating this estimation, the stresses also depend on the fluid speed and viscosity. Moreover, the stress measured by a particular sensor depends on the robot's orientation, even though the stress as a function of direction with respect to the vessel is independent of orientation for circular robots. We simplify the estimation procedure by removing these extraneous dependencies with the normalization described below.

\subsection{Fourier Coefficients}
\sectlabel{Fourier}

For the situations considered here, stresses vary fairly smoothly over the robot surface, as illustrated in \fig{3D and 2D}a. 
This observation motivates representing the stresses by a Fourier expansion in modes with low spatial frequencies.  Limiting consideration to these modes reduces the effect of noise (see \sectA{noise}), and is also a useful technique at larger sizes and Reynolds numbers~\cite{vollmayr14}.

The Fourier coefficient for mode $k$ is
\begin{equation}\eqlabel{Fourier}
\vec{c}_k = \frac{1}{n} \sum_{j=0}^{n-1} \vec{f}_j \exp\left(2\pi i \, \frac{j k}{n}\right)
\end{equation}
where $n$ is the number of sensors and $k$ ranges from 0 to $n-1$. 
To focus on low-frequency modes, we consider only modes up to $k=M$, with $M<n/2$. The examples in this paper use $M=6$ and $n=30$.

We use these coefficients in two ways.

First, we extend stress measurements from the locations of the robot's sensors to the full surface by interpolating the Fourier modes.
Mode $k$ is the real part of $\vec{c}_k e^{-i \theta k}$,  
where $\theta$ is the angle around the robot's surface measured counterclockwise from the front of the robot (see \fig{robot with sensors}b).
The interpolated stress values are
\begin{equation}\eqlabel{interpolation}
\vec{f}(\theta) = \Re\left(  \vec{c_0} +  2 \sum_{k=1}^M \vec{c_k} e^{-i \theta k}   \right)  
\end{equation}
where $\Re$ denotes the real part and the factor of 2 accounts for the symmetry between modes $k$ and $n-k$.

Second, we apply machine learning to determine relations between these Fourier coefficients and properties of the vessel near the robot.
To simplify this procedure,
we use the \emph{relative magnitudes} of the Fourier coefficients $\vec{m}_k$, with components
\begin{equation}\eqlabel{normalized Fourier}
m_{k,s} = \frac{1}{C} \left| c_{k,s} \right|
\end{equation}
where $s$ denotes either the normal or tangential component of the stress, and $C$ is the root-mean-square magnitude of the coefficients, i.e., $C=\sqrt{\sum_k  \left| \vec{c}_k \right|^2}$ with $\left| \vec{c}_k \right|^2 = \sum_s \left| c_{k,s} \right|^2$.
The constant mode, $\vec{c_0}$, does not contribute to the variation in stresses around the robot, so is not included in this normalization.
The relative magnitudes simplify the estimation because they are independent of the robot's orientation, the overall fluid flow speed and the fluid's viscosity. This is because 1) the robot's orientation affects the phase of the Fourier coefficients, but not their magnitudes, and 2) the stresses for Stokes flow are proportional to the flow speed and fluid viscosity~\cite{kim05}, 
so the normalization removes this dependence.

\subsection{Estimating the Direction of Motion}
\sectlabel{direction of motion}

As described in \sect{orientation}, a robot can estimate its direction of motion through the vessel from the observation, shown in 
\fig{3D and 2D}a, that the direction of motion is nearly perpendicular to the direction to the wall. Specifically, the figure shows the derivative, $d{f_\text{normal}}/d\theta$, of the normal stress at $\angleTExtreme$ is negative. If the direction of flow, and hence motion, were reversed, this derivative would be positive. Thus the sign of this derivative, $s$, identifies whether the motion is $+90^\circ$ or $-90^\circ$ from the direction to the wall, i.e., the direction of motion is $\angleToWall - s\, 90^\circ$.

\subsection{Relating Stresses to Robot and Vessel Properties}

Learning methods have a variety of trade-offs among expressiveness, accuracy, computational cost for training and use, and availability of training data~\cite{abu-mostafa12,jordan15,hastie09}. 
This paper focuses on regression methods, which perform well with only a modest number of training instances, and whose trained models require only a relatively small number of computations to evaluate.

\subsubsection{Principal Components of Fourier Coefficients}
\sectlabel{principal components}

To estimate the robot's position in the vessel and the vessel diameter, we use training samples to fit regressions between these properties and the Fourier coefficients of the stresses. However, these Fourier coefficients are highly correlated in these samples, which make regressions numerically unstable. To avoid this problem, we remove these correlations by using the main principal components of the Fourier coefficients. That is, we train regressions using just the principal components accounting for most of the variation among the modes~\cite{golub83}. 
With $M=6$ Fourier coefficients for each of normal and tangential stresses, the first two principal components account for $98\%$ of the variance  among the training samples. 
Thus we use just the first two principal components for our estimation.

Specifically, principal components are linear combinations of the relative magnitudes of the Fourier coefficients, $m_{k,s}$ given in \eq{normalized Fourier}. Specifically, the $h^{th}$ component has the form
\begin{equation}
p_h = \sum_{k=1}^M \sum_s a_{h,k,s} (m_{k,s} - \hat{m}_{k,s})
\end{equation}
where $\hat{m}_{k,s}$ is the average of $m_{k,s}$ over the training samples and $a_{h,k,s}$ are weights, with values for the first two principal components given in \tbl{principal components}.

\begin{table}
\centering
\begin{tabular}{ccccccc}
$k$		& 1 & 2 & 3 & 4 & 5 & 6 \\ \hline
$h=1$	& -0.419 & 0.508 & -0.268 & 0.060 & 0.027 & 0.010 \\
		& -0.419 & 0.473 & -0.298 & 0.047 & 0.018 & 0.006 \\ \hline
$h=2$	& 0.451 & 0.184 & -0.266 & -0.352 & -0.181 & -0.080 \\
		& 0.451 & 0.479 & -0.097 & -0.258 & -0.132 & -0.052 \\
\end{tabular}
\caption{Weights for first two principal components, i.e., $a_{h,k,s}$ for $h=1,2$. For each component, weights for normal and tangential stresses are on the first and second lines, respectively.}\tbllabel{principal components}
\end{table}

\begin{figure}
\centering \includegraphics[width=\figwidth]{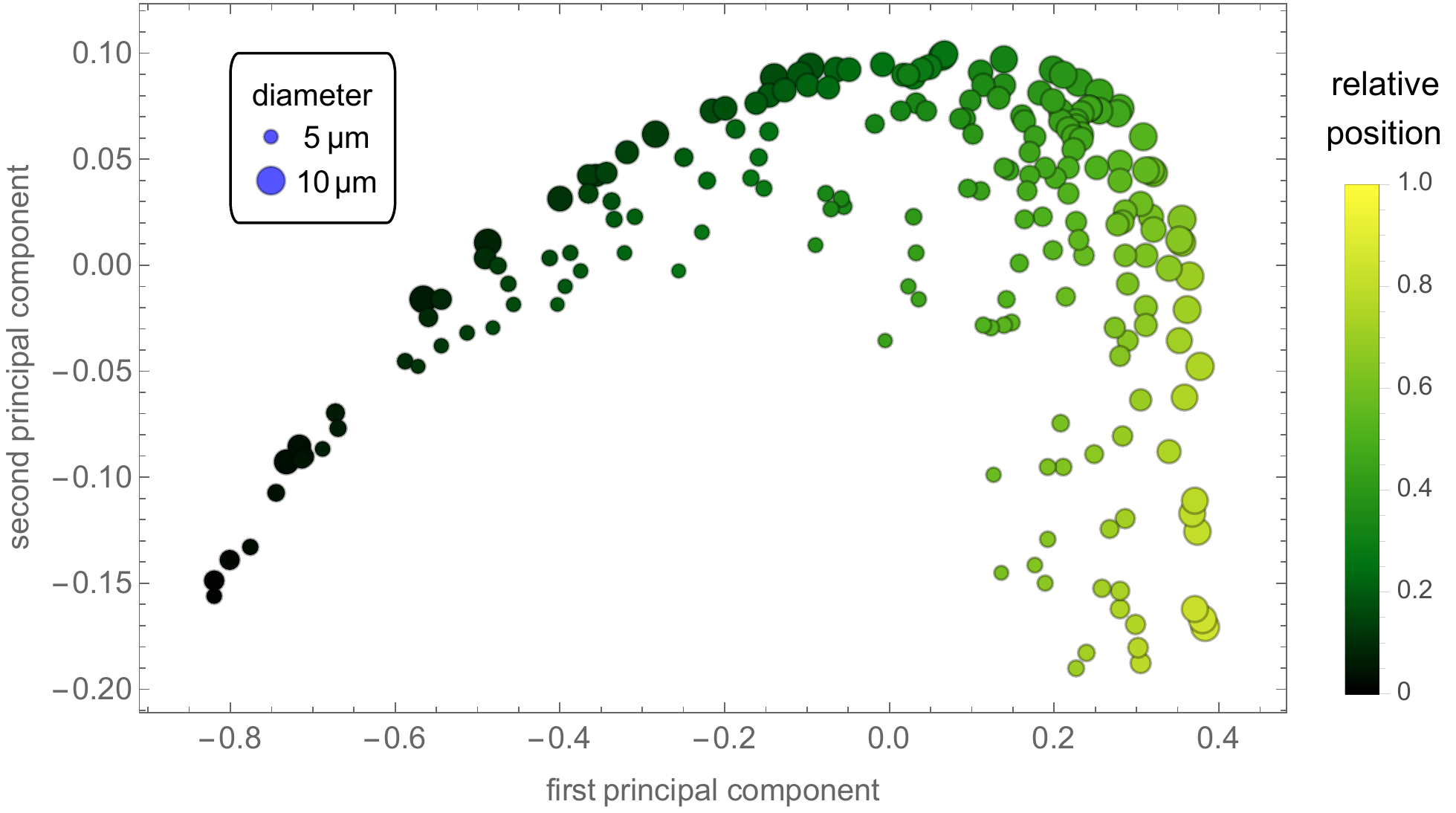}
\caption{First two principal components of the relative magnitudes of Fourier coefficients for 200 samples of a robot in a vessel. Disk sizes indicate vessel diameter and colors show the robot's relative position, given by \eq{relative position}.}\figlabel{principal components}
\end{figure}

\fig{principal components} indicates the range of the first two principal components for the samples described in \sect{samples}. The lower left portion of the figure has cases with the robot near the center of the vessel. The lower right has robots near the vessel wall. Thus, the first component mainly varies with the robot's position in the vessel and the second with the vessel's diameter.
The small spread in points for robots near the center of the vessel indicates relatively little information on vessel geometry from pattern of stress in that case. This leads to worse estimation performance when robots are near the center of the vessel.

\subsubsection{Regressions for Relative Position and Vessel Diameter} 
\sectlabel{regressions}

Because relative position is between zero and one, a convenient estimation method is a logistic regression. Using the training samples described in \sect{samples}, we relate relative position to the first two principal components of the Fourier coefficients,  $p_1$ and $p_2$, by
\begin{equation}\eqlabel{relative position regression}
\relPos = \frac{1}{1+\exp \left(- \left(\beta_0 + \beta_1 \, p_1 +  \beta_2 \, p_2 \right) \right)}
\end{equation}
where the $\beta_i$ are the fit parameters given in \tbl{relative position}.
\sect{relative position} evaluates the accuracy of this regression.

\begin{table}
\centering
\begin{tabular}{ccc}
parameter		& value	& standard error \\ \hline
$\beta_0$		& $-0.51$	& $0.09$ \\
$\beta_1$		& $3.3$	&  $0.3$ \\
$\beta_2$		& $-4.6$	&  $1.1$ \\
\end{tabular}
\caption{Regression parameters for relative position, \eq{relative position regression}.}\tbllabel{relative position}
\end{table}

Since diameter is a positive value, we create an estimator constrained to give positive values. Specifically, we use a generalized linear model~\cite{dobson08} with logarithm link function and gamma distribution for the residual. Vessel diameter has a strong nonlinear relation with the first two principal components of the Fourier coefficients, as seen in \fig{principal components}. A regression using only linear variation of the principal components gives poor estimates. Including quadratic terms accounts for much of the nonlinear relationship, giving an estimate for the diameter $d$, in microns, as
\begin{equation}\eqlabel{diameter regression}
d = \exp \left(\beta_0 + \sum_{i=1}^2 \beta_i \, p_i  + \sum_{1 \leq i \leq j \leq 2} \beta_{i,j} \, p_i p_j \right)
\end{equation}
with the parameters given in \tbl{diameter}.
\sect{diameter} evaluates the accuracy of this regression.

\begin{table}
\centering
\begin{tabular}{ccc}
parameter		& value	& standard error \\ \hline
$\beta_0$		& $1.66$	& $0.01$ \\
$\beta_1$		& $0.68$	&  $0.01$ \\
$\beta_2$		& $4.05$	&  $0.06$ \\
$\beta_{1,1}$		& $2.74$	&  $0.05$ \\
$\beta_{2,2}$		& $11.0$	&  $0.4$ \\
$\beta_{1,2}$		& $-5.4$	&  $0.2$ \\
\end{tabular}
\caption{Regression parameters for diameter, \eq{diameter regression}.}\tbllabel{diameter}
\end{table}

\subsection{Estimating Angular Velocity}
\sectlabel{estimating angular velocity}

As described in \sect{angular velocity}, a robot can use the shift in its surface stress pattern over a short time $\Delta t$ to estimate its angular velocity. This estimation requires the robot to have a clock with millisecond precision. 

Suppose $\vec{f}(\theta,t)$ is the stress at angle $\theta$ and time $t$. If the stress maintains its shape as the robot rotates, then $\Delta \theta$ would be the value such that $\vec{f}(\theta+\Delta \theta, t+\Delta t)=\vec{f}(\theta,t)$ for all $\theta$. However, any change in the shape of the stress pattern will mean there is no such value. A more robust approach to determine $\Delta \theta$ is the value maximizing the correlation between the stresses at the two times.
A check on the assumptions underlying this method is that the maximum correlation is close to one: indicating the main change in stresses is rotation around the robot, rather than, say, significant change in vessel geometry or the robot's distance to the vessel wall during time $\Delta t$.

This procedure must compare stresses close enough in time to avoid the robot completing a full turn, i.e., avoiding aliasing. For instance, the robot could measure the shift at a sequence of increasing intervals of time until there is a sufficient change in the angle of maximum correlation to reliably identify the change, but not so much that the shift exceeds a full turn.
A simpler approach is to use a fixed time interval. For the cases considered here, angular velocities are typically $100\,\radian/\second$, so $5\,\millisecond$ measurement interval gives a change in orientation of less than one radian, thereby avoiding aliasing.

One way to apply this method is to evaluate the correlation at shifts of an integer number of sensors. This is a simple computation, but limits $\Delta \theta$ to be an integer multiple of spacings between sensors. A more precise method, used here, has the robot interpolate its stress measurements (\eq{interpolation}), and determine $\Delta \theta$ as the value maximizing correlation between interpolated stress values at the two times. Estimates based on normal and tangential stresses are similar. As a specific choice, we use the average from these two stress components.

\subsection{Computational Requirements}
\sectlabel{computation}

Microscopic robots are likely to have limited on-board computational capabilities. Thus an important metric for the methods proposed here is their computational cost. 

The methods discussed in this paper use simple functions of a few Fourier coefficients of stress measurements. 
For $n$ sensors, a direct evaluation of $M$ Fourier coefficients uses of order $M n$ arithmetic operations\footnote{Faster methods are available for computing many modes, but are not necessary if $M$ is fairly small, as is the case considered here.}.
Normalizing the Fourier coefficients involves summing the magnitudes of the $M$ coefficients and then dividing these magnitudes by this sum.
Principal components are linear combinations of the normalized modes. Thus the arguments used in the regressions presented in this paper involve of order $M n$ multiplies and additions.
These evaluations involve a few thousand arithmetic operations for $n\approx 30$ sensors and $M\approx 6$ modes.

Evaluating stress patterns every, say, $10\,\millisecond$, would require less than $10^6\,\text{operations}/\second$.
This is comparable to the capabilities of microprocessors in the early 1980s, but in a much smaller volume. 
Such computation is well beyond the demonstrated capabilities of nano-scale computers, e.g., DNA-based logic in nanoparticles~\cite{douglas12}. However, in principle, molecular computers could be small enough for the robots discussed in this paper and readily exceed this computational performance~\cite{drexler92,merkle18}.

\section{Sensor Noise}
\sectlabel{noise}

In spite of ongoing progress in microscopic pressure sensors~\cite{cao17}, it remains a challenge to fabricate sensors for micron-scale devices that can detect the small stresses of the fluid flows considered here. An important practical question is the limitations of such sensors. This section discusses one such issue, the effects of thermal noise.

For definiteness, we consider a mechanical stress sensor on the surface of the robot. This could be a piston or membrane whose displacement changes under applied stress. One motivation for mechanical sensors converting force into displacement is their convenience for subsequent processing with mechanical computers~\cite{merkle18}.

\subsection{Noise for a Single Sensor}

We model a stress sensor as a stochastic damped oscillator with random variables $X$ and $V$ for its position and velocity, respectively. 
For definiteness, we take the equilibrium position $X=0$ when there is no applied stress.
To reduce noise, the robot averages measurements for time $t$. We evaluate the effect of this averaging with an additional random variable for the accumulated position measurements, $A(t) = \int_0^t X(t')\,dt'$. The time-averaged measurement is $A(t)/t$.

The behavior of this oscillator subject to thermal noise is given by a system of stochastic differential equations~\cite{gillespie92}:
\begin{equation}\eqlabel{noise}
\begin{split}
dV &= \left( -\gamma V - \omega^2 X + \alpha \right) dt + \sigma \,dW\\
dX &= V \,dt\\
dA &= X\,dt
\end{split}
\end{equation}
where $\gamma$ is the velocity damping coefficient, $\omega$ is the oscillator's natural frequency, $\alpha=f_{\text{applied}}/m$ is the applied force divided by the oscillator's mass $m$, and $dW$ is a standard Wiener stochastic process~\cite{gillespie92} representing thermal fluctuations of the oscillator's velocity. The fluctuation parameter $\sigma$ is the magnitude of the thermal noise.

An important characterization of the noise is the signal-to-noise ratio (\SNR), i.e., ratio of oscillator energies due to the applied force and thermal fluctuations. In terms of the stochastic model, \eq{noise}, \SNR\ equals the ratio of the squares of the mean and standard deviation of $A(t)/t$.
When the applied force, $f_{\text{applied}}$, changes slowly compared to oscillator damping time, measurements are described by the stochastic oscillator's equilibrium distribution. In that case, $A(t)/t$ is normally distributed with mean $\alpha/\omega^2$ and standard deviation $\sigma/(\omega^2 \sqrt{t})$~\cite{gillespie92}. 
Thus averaging measurements over times long compared to oscillator damping time but short compared to significant changes in the applied force gives
\begin{equation}\eqlabel{signal to noise}
\SNR =  \frac{\alpha^2 t}{\sigma^2}
\end{equation}

Relating this expression for \SNR\ to the robot's environment requires an estimate of $\sigma$. We obtain such an estimate by relating the noise magnitude to the oscillator damping, $\gamma$, via the fluctuation-dissipation theorem~\cite{gillespie92}: $\sigma = \sqrt{2 \BoltzmannConstant T \gamma/m}$ where $T$ is the temperature, $m$ the oscillator's mass, and $\BoltzmannConstant$ is Boltzmann's constant.

Damping of the sensor has two contributions. The first is from the fluid drag as the sensor surface moves through the fluid surrounding the robot. The second is from friction of the sensor's mechanisms inside the robot.
In principle, the internal friction can be made very small by using atomically-precise materials~\cite{cumings00,vanossi13}.
Thus, for this discussion, we assume the damping is primarily from fluid viscosity. 

An object of characteristic size $s$ moving at speed $v$ through a fluid with viscosity $\viscosity$ at low Reynolds number experiences a drag force $f_{\text{drag}} = g \viscosity s v$. The corresponding velocity damping coefficient is $\gamma = f_{\text{drag}}/(m v) =g \viscosity s/m$.
In these expressions, $g$ is a dimensionless geometric damping factor depending on the object's shape and orientation, as well as its relation to any nearby boundaries. 
For instance, in unbounded fluid, a sphere with diameter $s$ has $f_{\text{drag}} = 3 \pi \viscosity s v$~\cite{berg93}, so $g=3\pi$.
Similarly, a flat disk with diameter $s$ moving face-on through the fluid has $g = 8$~\cite{berg93}. 
These values provide rough estimates for the viscous damping of a stress sensor.

A sensor of size $s$ has surface area of about $s^2$. 
When subjected to stress $p$, the applied 
force per unit mass is $\alpha = p s^2/m$.
Combining these expressions in \eq{signal to noise} gives
\begin{equation}\eqlabel{signal to noise 2}
\SNR \approx \frac{p^2 s^3 t}{2 \BoltzmannConstant T g \viscosity}
\end{equation}

For a particular robot task, the fluid viscosity, temperature and magnitude of the stress, $p$, are fixed quantities. The design choices affecting noise are the sensor size $s$, the measurement averaging time $t$ and the sensor shape, which determines the damping factor $g$. \eq{signal to noise 2} shows that larger sensors, longer averaging times and smaller damping give larger signal-to-noise.

 \eq{signal to noise 2} allows comparing the effect of noise for different flow conditions.
For instance, stresses on a robot are proportional to the fluid speed and viscosity for a given vessel and robot geometry, i.e., $p \propto u \viscosity$.
Thus for fluid moving at a given speed, \eq{signal to noise 2} gives $\SNR \propto \viscosity$, so measurements are more accurate in more viscous fluids.
On the other hand, the effect of fluid speed on noise involves a trade-off. Slower speed gives smaller stresses, which are harder to detect. But slower robot motion also gives more time to average measurements before moving a given distance, e.g., a few times the robot's size. These changes scale as $p \propto u$ and  $t \propto 1/u$, respectively. Thus the decrease in stress dominates this trade-off, $\SNR \propto p^2 t \propto u$, so sensor noise limits stress-based navigation to sufficiently rapid flow speeds.

\subsection{Noise for Fourier Coefficients}

Stress-based estimates rely on combined measurements from $n$ sensors on the robot surface. Specifically, the methods described in this paper use the first few Fourier coefficients and their main principal components. These are weighted linear combinations of the sensor measurements with weights having similar magnitudes. Since thermal noise at different sensors is not correlated,  the noise associated with these combinations is about $1/\sqrt{n}$ times that of each sensor individually.

For $n$ sensors spread over a fraction $\surfaceFraction$ of robot's surface area, the sensor area is $s^2 = 4\pi r^2 \surfaceFraction/n$, where $r$ is the robot's radius. 
The signal to noise for the linear combinations is about $n$ times larger than that for the single sensor of \eq{signal to noise 2}, giving
\begin{equation}\eqlabel{signal to noise multiple}
\SNR_n \approx \frac{4 p^2 r^3 t}{\BoltzmannConstant T g \viscosity} \;  \sqrt{\frac{(\pi \lambda)^3}{n}}
\end{equation}

As a numerical example, consider the $n=30$ sensors discussed in \sectA{measurements} covering $\surfaceFraction=0.5$ of the robot's surface, with parameters of \tbl{fluid parameters}, typical stresses ($p\approx 1\,\pascal$) shown in \fig{3D and 2D}a, $g = 8$ (corresponding to a disk moving face-on), and averaging measurements for $t=5\,\millisecond$. In this case, $\SNR_n \approx 210$ or $23\,\decibel$.
The corresponding damping constant is $\gamma \approx 3.8\times 10^7/\second$, with damping time $1/\gamma \approx 30\,\nanosecond$, assuming the average density of the oscillator, of volume $s^3$, is close to that of water. This is much shorter than the millisecond measurement times considered here, justifying the assumption of equilibrium behavior for the stochastic oscillator used to derive \eq{signal to noise}.

Typically, a robot experiences larger stresses when closer to the vessel wall, making noise less significant. Combined with the higher accuracy of estimates in such cases (see \sect{geometry} and \sect{motion}), this means the methods considered here are most accurate for robots near the vessel wall.

A robot can determine that stresses are too weak to evaluate reliably due to noise if its principal component estimates are well outside the range of values indicated in \fig{principal components}. This would indicate the robot must average measurements over longer times for reliable estimates, or take action to increase stresses, such as moving toward the vessel wall.

\section{Generalizations}
\sectlabel{generalization}

\begin{figure}
\centering \includegraphics[width=\figwidth]{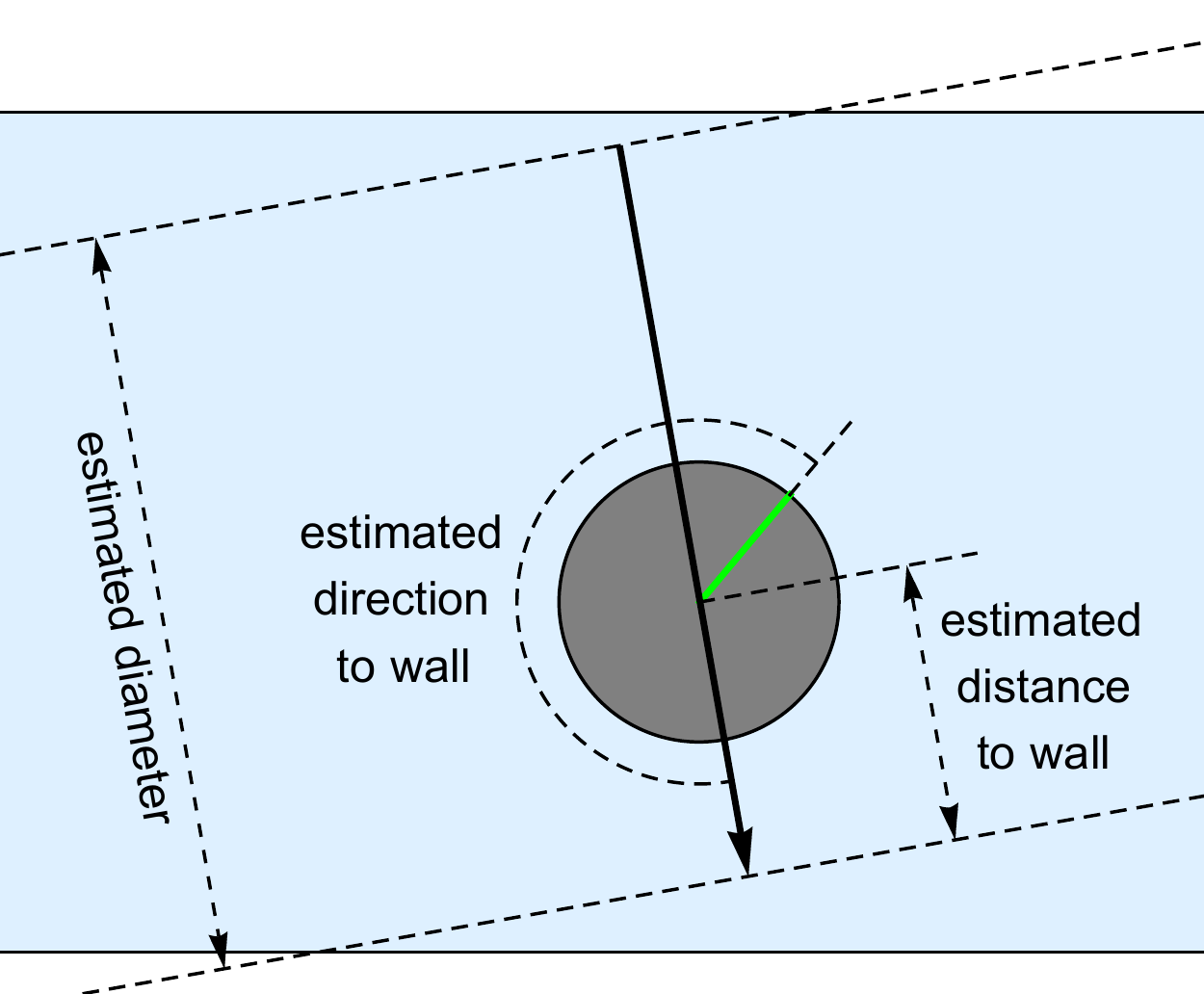}
\caption{Schematic illustrating estimates of a robot's relation to the vessel. The gray disk is the robot, and the green line shows its orientation. The solid arrow through the robot's center summarizes three estimates: the direction and distance from the robot's center to the nearest vessel wall, and the vessel's diameter. The shaded horizontal rectangle shows a portion of the actual vessel, with its top and bottom edges indicating the vessel walls. In this schematic, the robot slightly underestimates both its distance to the wall and the vessel's diameter. Perfect estimates would show the solid arrow directed perpendicular to the vessel walls and extending from one wall to the other.}\figlabel{estimates}
\end{figure}

The estimates described in this paper were developed for a single spherical robot moving passively with the fluid in a straight vessel. 
One approach to more complex situations is to create new estimates from samples of those cases. Unfortunately, this requires a large number of samples, e.g., for vessels with many shapes, and potentially more complex estimation procedures. An alternative, and simpler, approach is to apply the estimation methods of this paper to these situations, even though that will likely give less accurate estimates. 
This section illustrates this approach.
Specifically, we show the quality of the estimates of the robot's relation to the vessel with an arrow through the robot's center, as described in \fig{estimates}.

\subsection{Curved Vessels}
\sectlabel{curve}

\begin{figure}
\centering \includegraphics[width=\widefigwidth]{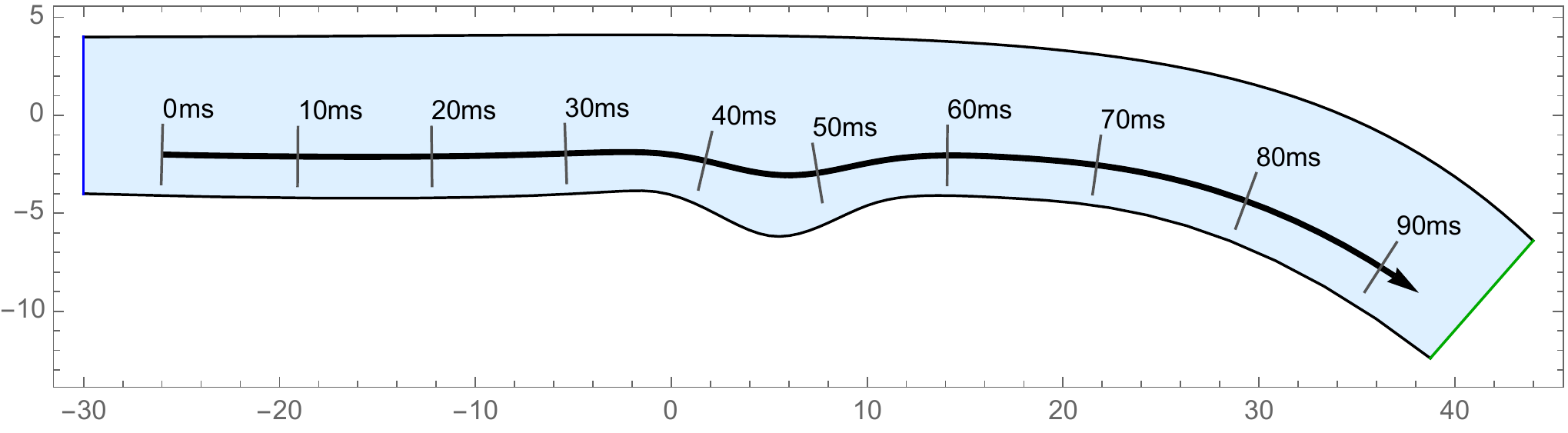}
\caption{A vessel with changing geometry. Positions along the axes are in microns. The curved arrow shows the path of the robot's center as fluid moves it through the vessel. Ticks along the curve indicate positions at various times. 
}\figlabel{curved vessel}
\end{figure}

This section examines how estimates based on straight vessels perform as a robot moves through a curved vessel. This vessel, shown in \fig{curved vessel}, has two types of curvature: a bump in the vessel wall with radius of curvature similar to the robot size, and gradual curvature, with radius of curvature large compared to the robot size. In this example, maximum inlet flow speed is $u=1000\,\micron/\second$.

\subsubsection{Estimating Angular Velocity}

\begin{figure}
\centering \includegraphics[width=\figwidth]{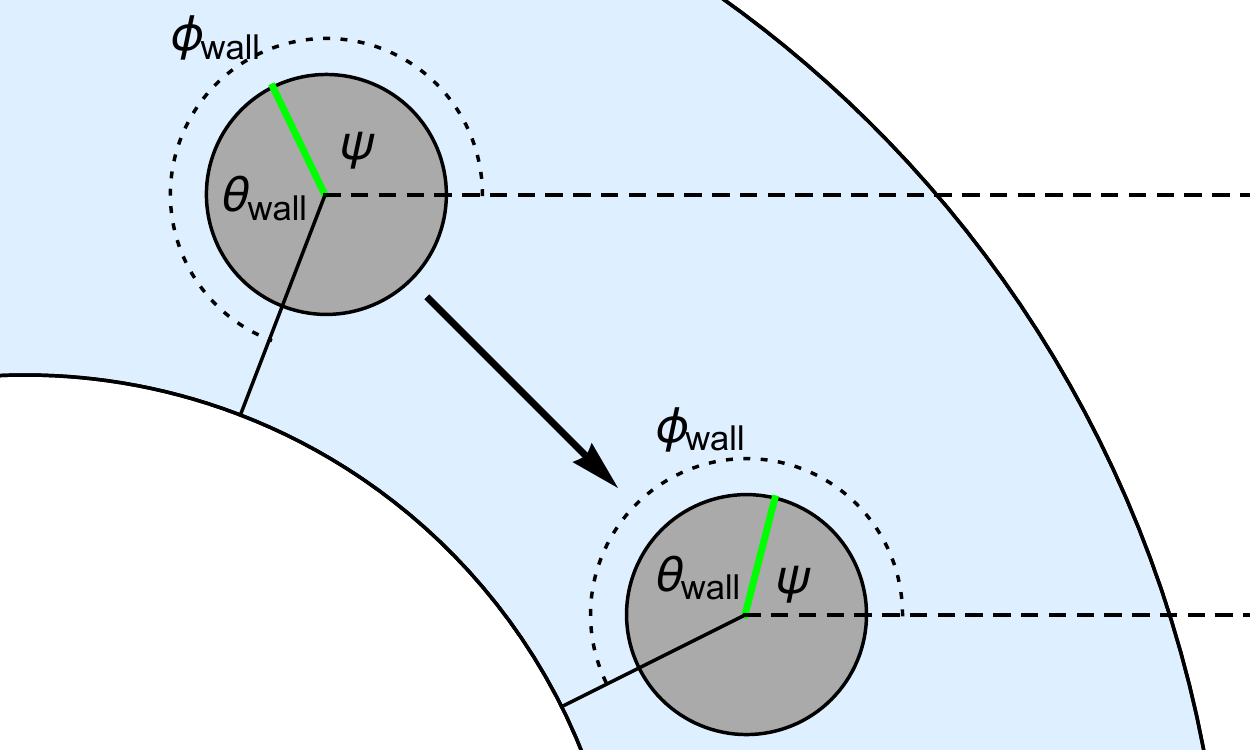}
\caption{Changing orientation and direction to the wall as a robot moves through a curved vessel.}
\figlabel{orientation schematic}
\end{figure}

As a robot moves through a vessel, the stresses on its surface mainly change due to the robot's rotation with respect to the vessel wall. In particular, the position on the surface of the maximum tangential stress generally faces the closest vessel wall. The direction to the wall changes due to a combination of the robot's rotation and the vessel curvature, as shown in \fig{orientation schematic}. Specifically,
\begin{equation}\eqlabel{changing direction to wall}
\frac{d\angleToWall}{d t} = \frac{d\angleXToWall}{d t} - \frac{d\orientation}{d t} 
\end{equation}
where $\orientation$ is the robot's orientation, and \angleToWall\ and \angleXToWall\ are the angle to the vessel wall measured from the front of the robot and from the $x$-axis, respectively, as illustrated in \fig{robot with sensors}b.
The first term on the right is due to the vessel curvature.
The second term is $-\omegaRobot$, the negative of the robot's angular velocity.

In general, measuring the change in stresses, as described in \sect{angular velocity}, estimates how rapidly the direction to the wall changes with respect to the front of the robot. Alternatively, $-d\angleToWall/dt$ is how rapidly the robot rotates with respect to the vessel wall, giving the rate of change of the orientation with respect to the wall described in \sect{orientation}. In a straight vessel, $-d\angleToWall/dt = \omegaRobot$ so the rate of stress change estimates the robot's angular velocity. However, in a curved vessel, \eq{changing direction to wall} shows this estimate has an additional contribution from the vessel curvature.

\begin{figure}
\centering 
\begin{tabular}{cc}
\includegraphics[width=\mfigwidth]{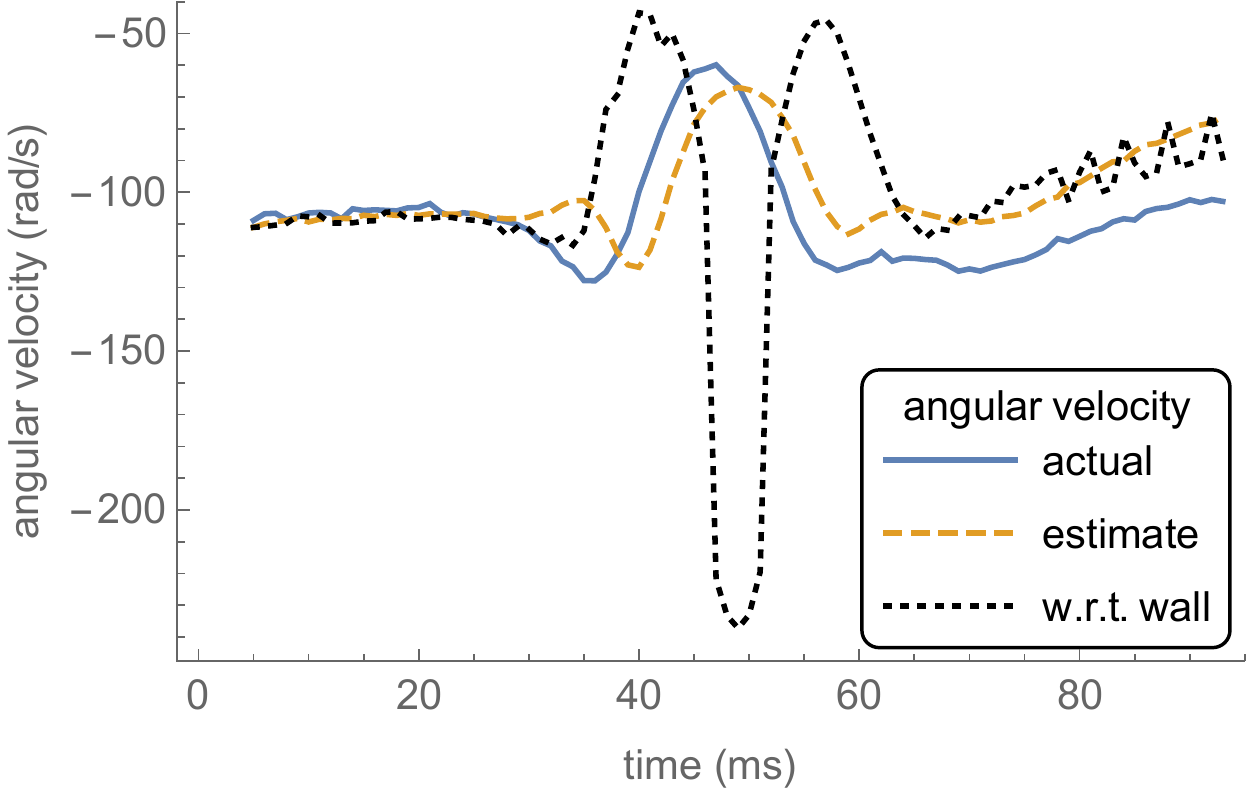}	&  \includegraphics[width=\mfigwidth]{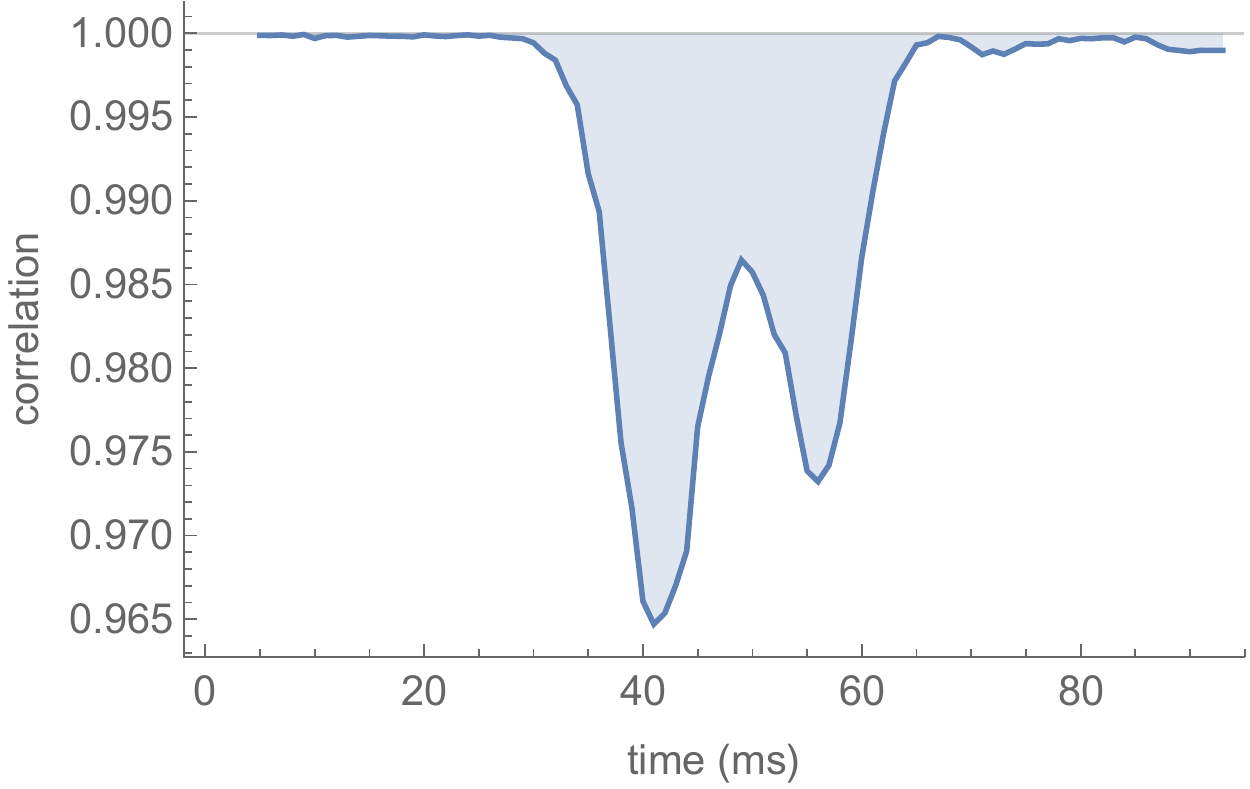} \\
(a)	&	(b) \\
\end{tabular}
\caption{Estimated angular velocity as robot moves through the vessel of \fig{curved vessel} based on stress correlation over $\Delta t=5\,\millisecond$. 
(a) Actual and estimated angular velocities, and the rate of change of the robot's orientation with respect to the wall.
(b) Correlation of interpolated stress patterns separated by $5\,\millisecond$.
}\figlabel{curved vessel angular}
\end{figure}

For a robot moving through the curved vessel, \fig{curved vessel angular}a compares the angular velocity, estimated from correlation over $\Delta t = 5\,\millisecond$ as described in \sect{angular velocity}, with the actual value, \omegaRobot. While the robot is in the straight portion of the vessel, the estimate is close to the actual value. The angular velocity changes as the robot passes the bump in the vessel wall, and the estimate tracks the change with a delay of about $\Delta t$. 

In the slowly curving portion of the vessel, the estimate deviates systematically from the actual value. This difference is because the estimation procedure actually measures the rate the robot's orientation changes with respect to the vessel wall, i.e., $-d\angleToWall/d t$ from \eq{changing direction to wall}.
To test this relationship, the figure also shows the average rate of change of the robot's orientation with respect to the wall. This is close to the estimate in both the straight and gently curved sections of the vessel. As the robot passes the bump, the estimates deviates significantly from $-d\angleToWall/d t$, in part because the direction of maximum tangential stress deviates from the direction to the nearest point on the vessel wall, as seen in \fig{curved vessel behavior}b. 
So in this case, large tangential stress is not an accurate indication of the direction to the vessel wall, and its change over $\Delta t$ also gives a poor estimate for $-d\angleToWall/d t$. Instead, for small, high-curvature bumps in the vessel wall, the estimate more closely tracks the actual angular velocity.

\fig{curved vessel angular}b shows the correlation between stress patterns separated by $5\,\millisecond$, the time interval used to estimate the angular velocity. The drop in correlation as the robot passes the bump indicates the stress pattern is changing more than it would just due to robot rotation in a straight vessel. The decreased correlation is an indication of abrupt geometry changes, and when the angular velocity estimates may be unreliable.

\subsubsection{Estimating the Robot's Relation to the Vessel}

\begin{figure}
\centering 
\includegraphics[width=\widefigwidth]{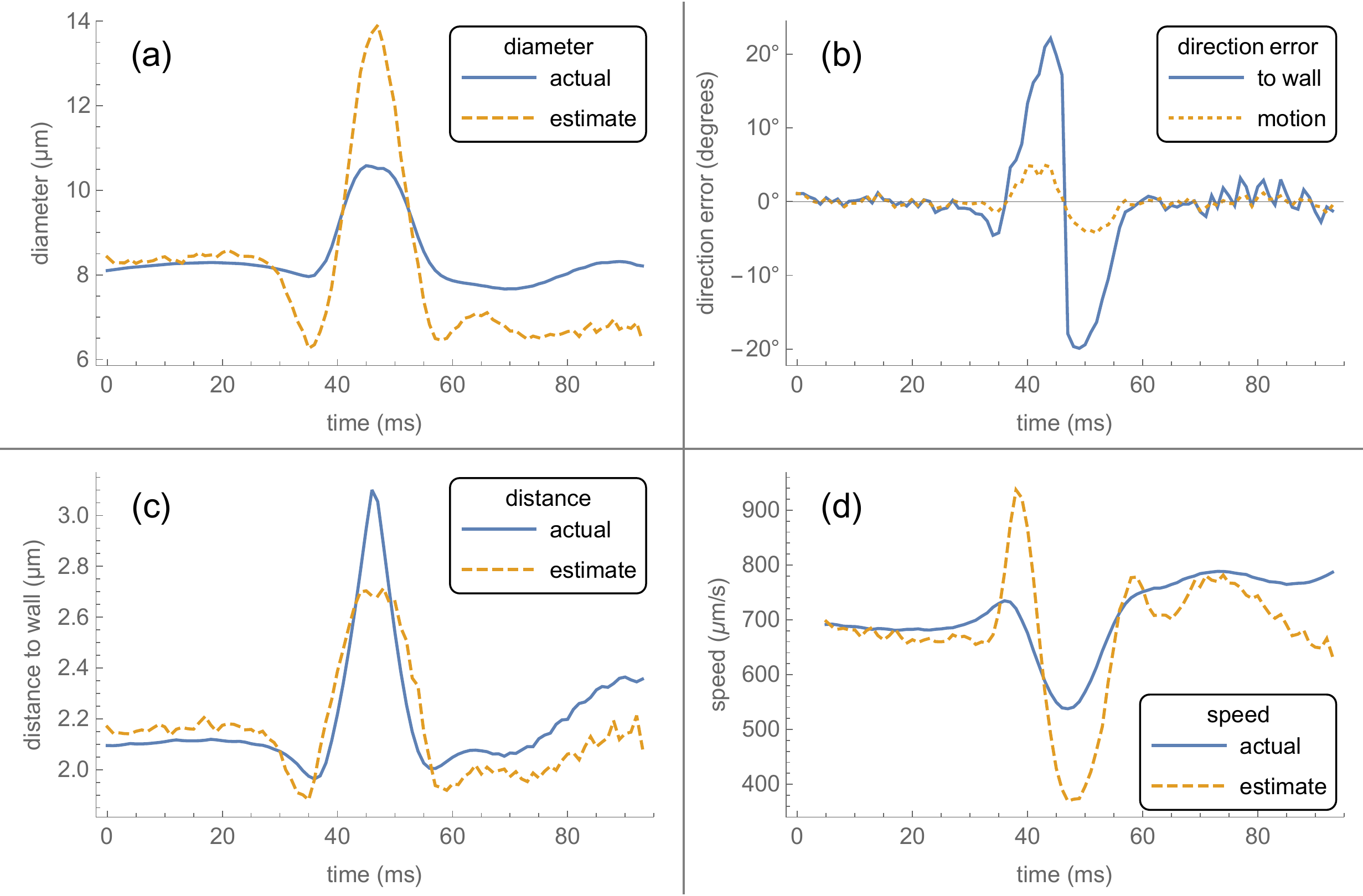}	
\caption{Comparison of actual and estimated values as a function of time as the robot moves through the vessel shown in \fig{curved vessel}. 
(a) Vessel diameter at the robot's position.
(b) Difference between estimated and actual directions to the nearest point on the wall and the direction of motion.
(c) Distance between the robot's center and the nearest point on the vessel wall.
(d) Robot's speed.
}\figlabel{curved vessel behavior}
\end{figure}

\fig{curved vessel behavior} compares actual and estimated values as the robot moves through the vessel. The robot is in a nearly straight vessel for the first $30\,\millisecond$, and the estimates are close to the actual values.
During $35\mbox{--}55\,\millisecond$, the robot passes the small bump in the vessel wall, and estimates deviate from the actual values. The fairly abrupt changes in the estimates allow the robot to identify when such changes in geometry occur, and thereby indicate when the estimates are likely to be less reliable.
Finally, after about $60\,\millisecond$, the robot moves in the gently curved section of the vessel. The direction estimates are quite accurate in spite of the curvature (\fig{curved vessel behavior}b) but the other estimates have somewhat larger errors than in a straight vessel.
Overall, the estimates give the robot a rough idea of its relation to the curved vessel.

\subsection{Elongated Robots}
\sectlabel{elongated}

This paper focuses on spherical robots, where stresses do not depend on the robot's orientation. However, elongated shapes could be useful for detecting chemical gradients~\cite{dusenbery09} and for locomotion~\cite{hogg14}. The pattern of stress on an elongated robot changes as it rotates, alternating between relatively long periods with its long axis nearly aligned with the flow and short flips.

\begin{figure}
\centering \includegraphics[width=\widefigwidth]{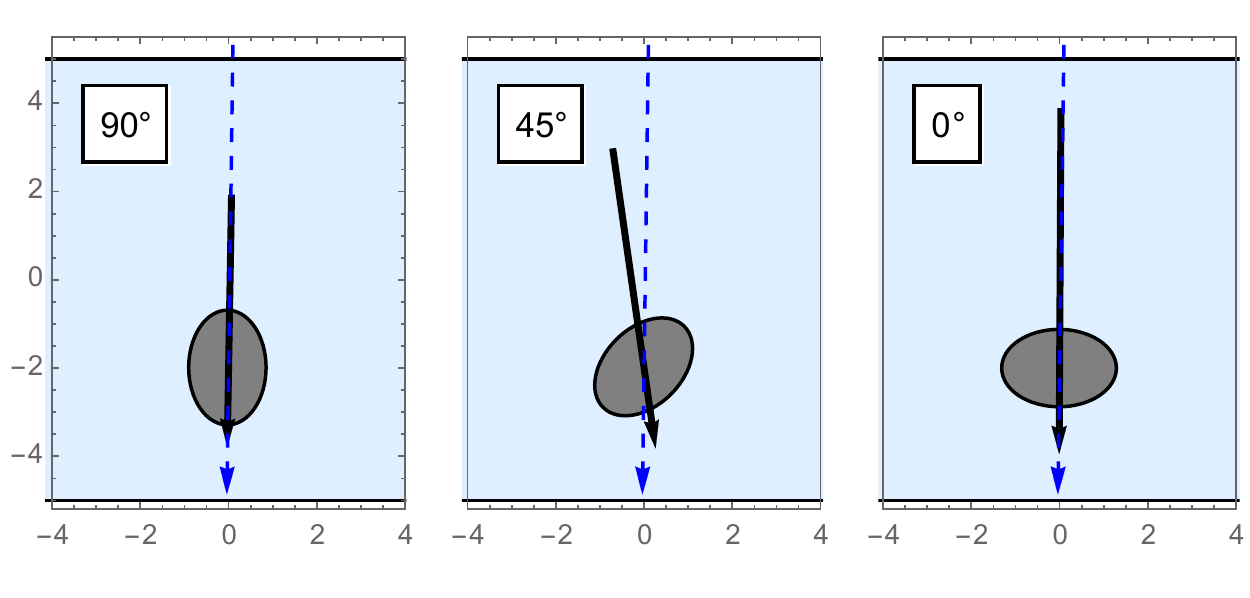}
\caption{Example of elongated robots at different orientations. Positions along the axes are in microns. Flow is from left to right. The angles in the upper left corners of the plots are the orientation of the robot, whose front is taken to be along the semi-major axis.
Solid arrows through each robot indicate the robot's estimates, as described in \fig{estimates}. The dashed blue arrows show the estimates for a circular robot with $1\,\micron$ radius at the same position. 
}\figlabel{ellipse}
\end{figure}

\begin{table}
\centering
\begin{tabular}{lccc}
shape	& orientation	& speed		& angular velocity\\
\hline
ellipse	& $90^\circ$	& $-32\%$	& $-0.7\%$\\
ellipse	&$45^\circ$	& $-44\%$		& $-1.6\%$\\
ellipse	&$0^\circ$		& $-65\%$		& $-3.3\%$\\
circle		& all values		& $-2.0\%$	& $-3.0\%$\\
\end{tabular}
\caption{Relative errors in the estimates of robot speed and angular velocity for the robots shown in \fig{ellipse} and a circular robot at the same position. The estimates are from the methods of \sect{motion}, based on a circular robot. The angular velocity estimate is from motion over $\Delta t=5\,\millisecond$ leading to the positions in \fig{ellipse}.}\tbllabel{ellipse}
\end{table}

Applying the methods developed here for elongated robots requires generalizing the interpretation of quantities involved in the estimation procedures, which use normal and tangential components of the surface stress. For a sphere, these components not only describe the relation of the stress vector to the surface, but also correspond to forces directed toward and perpendicular to the center of the robot. Only the latter apply torque around the center.
For an elongated shape, these are different decompositions of the stress, e.g., stress normal to the surface is not always directed toward the center and so can contribute to torque. Either of these vector decompositions, or other choices, may provide useful generalizations for elongated robots. For definiteness, we use the normal and tangential components of the stress on elongated robots.

As an example, consider a prolate spheroid with same volume as the spherical robots discussed in this paper, with semi-major axis $a=1.3\,\micron$ and semi-minor axis $b=r^{3/2}/\sqrt{a}$ where $r=1\,\micron$ is the radius of the spherical robot. 
The aspect ratio of this spheroid is $a/b=1.5$.

\fig{ellipse} shows how the estimates, using the methods developed for spherical robots, perform for this elongated shape at various orientations. In these diagrams, the estimated relative position gives the position of the robot's center, $y_c$, via \eq{relative position}, by generalizing the radius of the spherical robot to be the minimum distance between the robot's center and the wall in the estimated direction to the wall, i.e., the distance at which the robot would just touch the wall.

\subsection{Vessels Containing Multiple Objects}

In the above scenarios, the fluid moves a single robot through the vessel. In general, vessels could contain additional objects, such as other robots and blood cells. This section illustrates how the methods described above, developed for a single robot far from any other objects in the vessel, apply when other objects are nearby.

\subsubsection{Vessels With Cells Near the Robot}

\begin{figure}
\centering \includegraphics[width=\widefigwidth]{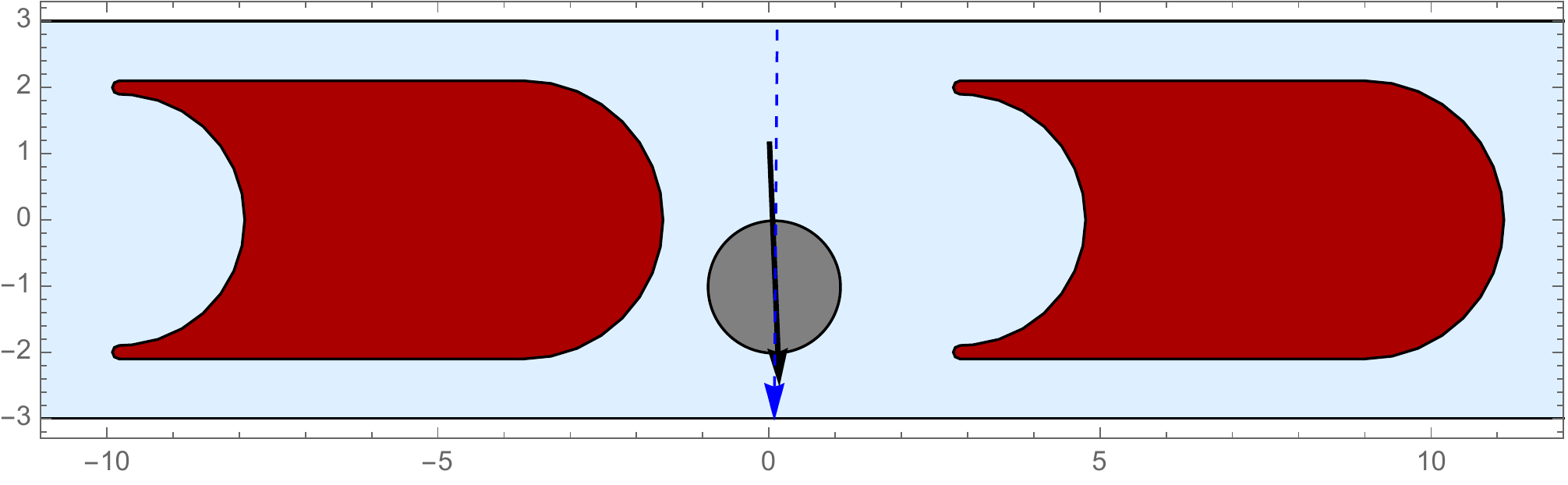}
\caption{Example of robot between two distorted cells in a small vessel. Positions along the axes are in microns. Flow is from left to right.
The solid arrow through the robot indicates it's estimates, as described in \fig{estimates}. The dashed blue arrow shows the estimates for the same robot in a vessel without the cells.
}\figlabel{cells in vessel}
\end{figure}

Blood cells deform as they enter capillaries, and move through in single file. In such cases, robots will be in fluid between two cells. \fig{cells in vessel} is an example with a robot between two deformed blood cells. Specifically, this model~\cite{secomb01} includes the distortion of cells as they enter small vessels, including the resulting gap between the cells and the vessel wall, but thereafter treats them as rigid bodies in their equilibrium shape, which can take a few tens of milliseconds to reach~\cite{secomb01}. 
In particular, this neglects any additional deformation of the cells due to changes in the flow from the rigid robot between them.
The geometry used here is the same as in a study evaluating the effect of cells on chemical sensors~\cite{hogg08h}, where
cells move at $1000\,\micron/s$ through a vessel with $6\,\micron$ diameter. The cells are cross sections of a 3D axially symmetric model of cells distorted into bullet-like shapes as they pass single-file through capillaries. These cells have volume $90\,\micron^3$ and surface area $135\,\micron^2$, which are typical values for red blood cells.
The spacing between the cells corresponds to $25\%$ hematocrit, a typical value for capillaries~\cite{freitas99}.

\begin{table}
\centering
\begin{tabular}{lcc}
			& speed	& angular velocity \\
\hline
with cells		& $42\%$		& $4.9\%$	\\
no cells		&$15\%$		& $-1.3\%$\\
\end{tabular}
\caption{Relative errors in the estimates of robot speed and angular velocity for the robot shown in \fig{cells in vessel} and a robot at the same position in a vessel without the cells. The estimates are from the methods of \sect{motion}, based on a robot in an otherwise empty vessel. The angular velocity estimate is from motion over $\Delta t=5\,\millisecond$ leading to the position in \fig{cells in vessel}.}\tbllabel{cells in vessel}
\end{table}

We estimate the robot's relation to the vessel using the methods of this paper, which were trained on a robot in an otherwise empty vessel. 
The solid arrow through the robot in \fig{cells in vessel} shows that the robot correctly estimates the direction to the vessel wall but significantly underestimates the vessel diameter.
\tbl{cells in vessel} compares actual and estimated robot motion for the robot. Angular velocity is estimated well, but speed is significantly overestimated.
In this case, the fluid moves the robot a bit faster than the cells, so the robot moves closer to the downstream cell. The robot also moves toward the center of the vessel. As the robot moves with respect to the cells, the errors in the estimates are similar to those for the specific case in \fig{cells in vessel}.  
Thus the presence of cells near the robot significantly affects the accuracy of the estimates. Nevertheless, the estimates remain within a factor of two of the actual values, thereby giving rough estimates without the need to develop new estimators for this geometry.

\subsubsection{Vessels With Multiple Robots}

In some applications, robots could operate in close proximity. Or, occasionally pass each other, e.g., as a robot near the center of a vessel overtakes slower robots near the wall.

As one example, \fig{multiple} shows the estimates for four robots near each other in a vessel. \tbl{multiple} shows relative errors in the estimates of speed and angular velocity.
Robots near the vessel wall have similar estimates with or without the other robots.
However, the robot in the middle of the group (i.e., number 3 in the figure) has significant errors in these estimates.
This robot could tell that it has poor estimates: those estimates indicate the robot is moving quickly in a narrow vessel. Thus the robot should encounter large stresses, which are about 40 times larger than the measured values in this case. Such small measured stresses are not plausible for two reasons. First, the small stresses would require the fluid speed or viscosity to decrease by that amount compared to prior measurements, before joining the group. Second, the robot could compare its measurements with the stresses communicated from the other robots, which are not consistent with high speed in a narrow vessel.

\begin{figure}
\centering \includegraphics[width=\smallfigwidth]{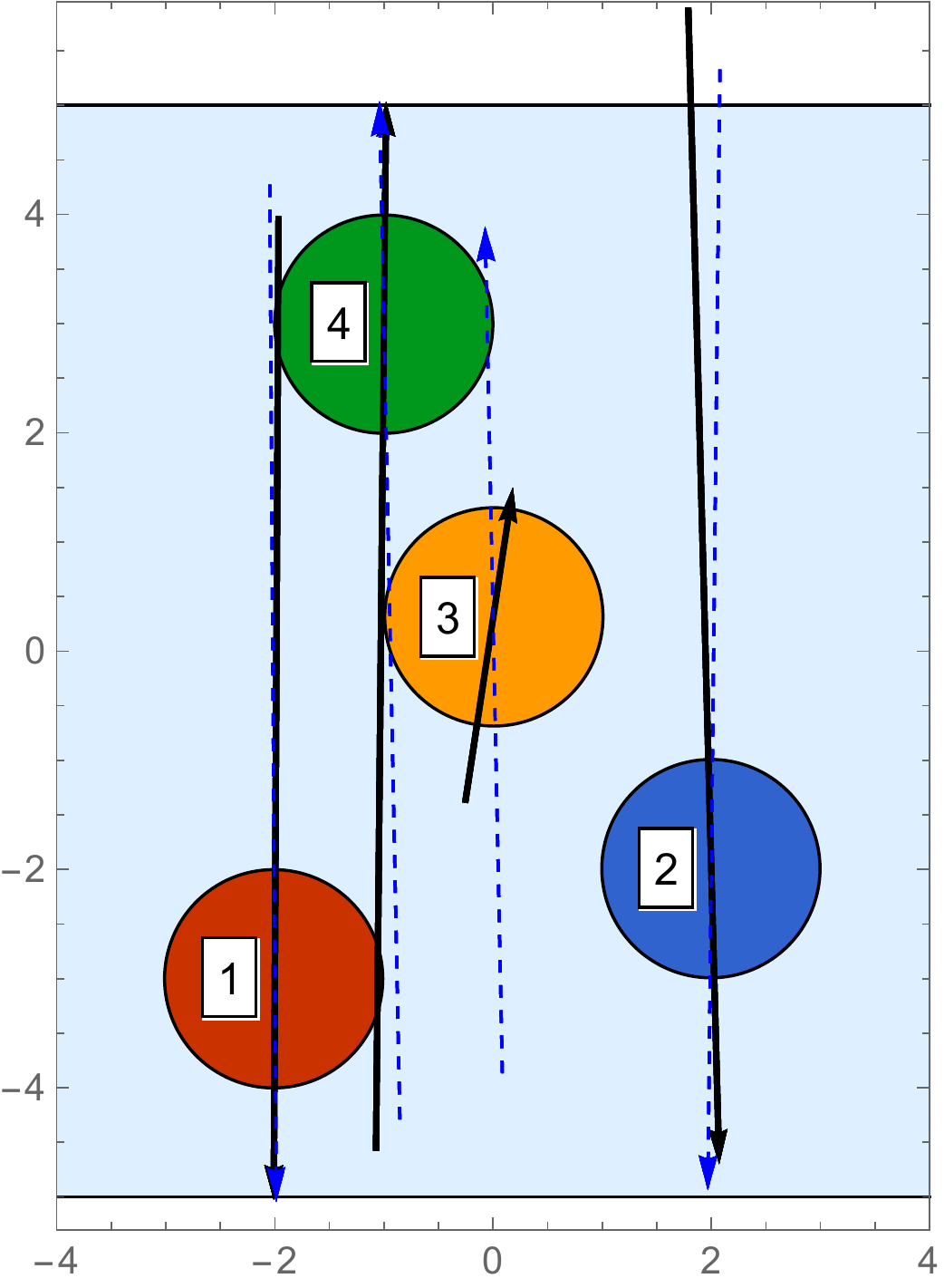}
\caption{Example of four robots in a small vessel. Positions along the axes are in microns. Flow is from left to right.
The solid arrows through each robot indicate that robot's estimates, as described in \fig{estimates}. Dashed blue arrows are the corresponding estimates for each robot at the indicated position but by itself in the vessel. The numbers label the robots from bottom to top in the vessel.
}\figlabel{multiple}
\end{figure}

\begin{table}
\centering
\begin{tabular}{ccccc}
			& \multicolumn{2}{c}{speed} 	& \multicolumn{2}{c}{angular velocity} \\
robot			& group	& by itself		& group	& by itself\\
\hline
1		& $-7.7\%$	& $-0.04\%$	& $3.6\%$		&$-0.7\%$\\
2		&$-35\%$		& $-24\%$		& $5.5\%$		&$-1.6\%$\\
3		&$130\%$		& $-40\%$		& $-1050\%$	&$-3.3\%$\\
4		&$-6.5\%$		& $-2.0\%$	& $6.2\%$		&$-3.0\%$\\
\end{tabular}
\caption{Relative errors in estimates of robot speed and angular velocity for each robot in the group shown in \fig{multiple} and when that robot is by itself, at the same position, in the vessel. The estimates are from the methods of \sect{motion}, based on a robot in an otherwise empty vessel. The angular velocity estimate is from motion over $\Delta t=5\,\millisecond$ leading to the positions in \fig{multiple}.}\tbllabel{multiple}
\end{table}

The robots could communicate to share estimates, allowing the group to select the most reliable one, or combine estimates. E.g., giving more weight to robots that estimate they are close to the wall, where the estimation methods discussed here are more reliable. Or more weight to robots with higher correlation in their measurements over time.

A caveat from this example is that stress-based estimates can not coordinate detailed motion of nearby robots. This is because precise coordination would take place over times during which robots move less than a body diameter, i.e., somewhat less than a millisecond, which is too short for averaging to reduce sensor noise enough for accurate estimates (see \sectA{noise}).

\subsection{Locomotion}

The above discussion considered robots passively pushed through the vessel by the moving fluid. Robots with locomotion capability can actively move through the fluid. Such locomotion can improve performance, e.g., using chemical propulsion~\cite{avila17} to improve \invivo\ drug delivery.
Locomotion alters the stresses on the robot surface compared to passive motion with the fluid. This section discusses methods to accommodate these changes in stress-based estimates.

One approach for locomoting robots is to directly use the stress patterns they encounter. This would involve training new estimators based on stress patterns that occur when the robot activates its locomotion with various speeds and patterns (e.g., faster motion on one side of robot to change its orientation). The robot would select among these trained estimators based on the current state of its locomotion. This state could be specified either by 1) how rapidly it moves locomotion actuators (e.g., treadmills), or 2) how much force or power it applies to locomotion actuators; whichever is easier for robot to specify or measure. While this approach is likely to be the most accurate, it requires extensive additional training over the range of locomotion actuations the robot will use. 

Another approach to handling locomotion is to vary the locomotion speed in such a way that the robot can continue to use estimators based on passive motion in the fluid.
Two methods for this are
\begin{itemize}
\item Occasionally turn off locomotion. The fluid quickly reverts to passive flow (e.g., in nanoseconds for micron-sized objects, such as robots and bacteria~\cite{purcell77}). Thus, in brief periods with no locomotion, the robot measures stress corresponding to passive motion.
The measurement time required with no locomotion depends on the stress magnitudes and sensor noise (see \sectA{noise}). This method is most appropriate when this time is short compared to significant passive drift by the robot while its locomotion is off.

\item Exploit the linear relation between speeds and stresses in Stokes flow~\cite{kim05} 
when the geometry does not change. For example, the robot could measure stresses with locomotion at twice and half the intended speed, and linearly extrapolate the stresses to zero locomotion speed. This gives the robot the stresses it would measure if it were moving passively with the fluid, without needing to turn off locomotion completely. The accuracy of this approach requires negligible change in geometry during this procedure, both for the robot and its environment. In particular, this requires that the locomotion itself does not change robot geometry significantly. E.g., the robot could use surface-based motions such as treadmills or small surface oscillations~\cite{hogg14}, rather than moving extended structures such as flagella. This is an example of selecting a robot hardware design to simplify control software, reducing computational requirements for the robot's computer, rather than focusing just on the performance of one robot component in isolation, e.g., the power efficiency of the locomotion method.
\end{itemize}
With both these methods, the robot obtains stress values corresponding to passive motion and so can apply the estimators developed in this paper for such motion. This approach requires no additional training with locomoting robots. On the other hand, the required changes to the intended locomotion, needed to measure or estimate stresses from passive motion, may reduce locomotion performance. E.g., the robot may require additional time to compensate for motion in undesired directions during the changed locomotion.


\clearpage

\end{document}